\documentclass{article}
\usepackage{arxiv}
\usepackage[T1]{fontenc}
\usepackage[utf8]{inputenc}
\usepackage{microtype}

\usepackage{amsmath,amsfonts,bm}

\def\eqref#1{equation~\ref{#1}}

\def\1{\bm{1}}

\DeclareMathAlphabet{\mathsfit}{\encodingdefault}{\sfdefault}{m}{sl}
\SetMathAlphabet{\mathsfit}{bold}{\encodingdefault}{\sfdefault}{bx}{n}

\usepackage[authoryear,round]{natbib}
\setcitestyle{authoryear,round,citesep={;},aysep={,},yysep={;}}
\usepackage{url}
\usepackage{multirow}
\usepackage{enumitem}
\setlist[itemize]{labelindent=0.8em,leftmargin=*,nosep}
\usepackage{soul}
\usepackage{graphicx}
\usepackage{algorithm}
\usepackage{algorithmic}
\usepackage{makecell}
\usepackage[table]{xcolor}
\usepackage{wrapfig}
\usepackage{subcaption}
\usepackage{booktabs}
\usepackage[hidelinks]{hyperref}
\hypersetup{
  pdftitle={Adamas: Hadamard Sparse Attention for Efficient Long-context Inference},
  pdfauthor={Siyuan Yan, Guo-qing Jiang, Yucheng Zhang, Xiaoxing Ma, Ran Zhu, Chun Cao, Jingwei Xu}
}

\title{Adamas: H\underline{adama}rd \underline{S}parse Attention\\
    for Efficient Long-context Inference}
\renewcommand{\shorttitle}{Adamas: Hadamard Sparse Attention}
\author{%
\textbf{Siyuan Yan\textsuperscript{1,2}\thanks{Equal Contribution. This work was partially done during Siyuan's internship at rednote.}\quad
Guo-qing Jiang\textsuperscript{2,*}\quad
Yucheng Zhang\textsuperscript{1}}\\
\textbf{Xiaoxing Ma\textsuperscript{1}\quad
Ran Zhu\textsuperscript{2}\quad
Chun Cao\textsuperscript{1}\quad
Jingwei Xu\textsuperscript{1}\thanks{Corresponding Author.}}\\
\textsuperscript{1}State Key Laboratory for Novel Software Technology, Nanjing University, China\\
\textsuperscript{2}rednote hilab, China\\
\texttt{siyuanyan@smail.nju.edu.cn, jingweix@nju.edu.cn}%
}
\date{}

\begin{document}
\maketitle

\begin{abstract}
Large language models (LLMs) now support context windows of hundreds of thousands to millions of tokens, enabling applications such as long-document summarization, large-scale code synthesis, multi-document question answering and persistent multi-turn dialogue. However, such extended contexts exacerbate the quadratic cost of self-attention, leading to severe latency in autoregressive decoding. Existing sparse attention methods reduce these costs by attending to only a subset of key-value (KV) tokens, but fixed or coarse-grained selection strategies can miss important tokens or retain substantial redundancy under constrained token budgets. We introduce \textbf{Adamas}, an accurate and efficient token-level sparse attention mechanism for long-context inference. Adamas applies the Hadamard transform, bucketization and 2-bit compression to produce compact representations, and uses lightweight distance-based estimation for efficient top-$k$ candidate selection. The selected positions are then used to gather the corresponding original KV states for sparse attention. Comprehensive evaluations show that Adamas maintains near-lossless performance within the effective context length of Llama-3.1-8B-Instruct. In efficiency evaluations, Adamas achieves up to $6.5\times$ self-attention speedup and $1.7\times$ end-to-end decoding speedup at a 32K context length. Adamas also maintains performance comparable to full attention on challenging long-form reasoning benchmarks, including AIME 2024 and AIME 2025.
\par\noindent
\textbf{Code:} \url{https://github.com/FibonaccciYan/Adamas}
\end{abstract}

\section{Introduction}
\label{sec:introduction}


The rapid progress of LLMs has dramatically extended their affordable context windows. Contemporary systems such as Anthropic’s Claude Sonnet 4 support up to 1M tokens~\citep{anthropic2025claude4sonnet}, while OpenAI’s GPT-5 effectively handles contexts up to 400K tokens~\citep{openai2025gpt5}. These expanded capacities enable advanced applications, including long-document summarization~\citep{chang2024booookscore}, large-scale code synthesis~\citep{nijkamp2023codegen, dainese2024generating}, multi-document question answering~\citep{wu2025mmqa, wang-etal-2024-leave}, and persistent multi-turn dialogue~\citep{collabllm2025}. By processing extensive information without manual segmentation, LLMs are increasingly able to tackle tasks that demand global coherence and long-term memory.
However, the quadratic complexity of self-attention~\citep{zaheer2020big, vaswani2017attention}, together with the growing KV cache, introduces substantial per-token latency and memory overhead during inference. 

Sparse attention offers a promising solution by restricting each query to attend only a carefully selected subset of tokens~\citep{yuan-etal-2025-native}.
This reduces the number of KV pairs involved in attention computation, lowering both computational complexity and memory access costs while largely preserving modeling capacity. Nevertheless, existing approaches often fall into two categories with notable limitations. Static sparsity patterns (e.g., StreamingLLM~\citep{xiao2023efficient}) are often hand-designed from empirical attention heatmaps, yielding fixed patterns, such as fixed local windows or vertical stripes. However, these patterns fail to capture the dynamic nature of query–key interactions, leading to low recall and degraded accuracy (See Table~\ref{tab:longbench_avg}). In contrast, dynamic methods like Quest~\citep{tang2024quest} adapt token selection during inference, but their page-level granularity remains overly coarse. This coarse selection introduces token redundancy and limits the achievable sparsity ratio, since higher sparsity levels lead to accuracy degradation. 

In this paper, we propose \textbf{Adamas}, an accurate and efficient token-level sparse attention mechanism for long-context inference. Adamas uses compact query and key representations to efficiently identify candidate KV tokens for each decoding query. Specifically, it first applies the Hadamard transform to queries and keys, which redistributes large-magnitude outliers and produces smoother feature distributions for coarse discretization. The transformed representations are then bucketized into compact 2-bit codes using calibrated thresholds and stored alongside the original KV cache with small additional memory overhead. During decoding, Adamas evaluates a lightweight distance-based score over the compressed representations to select the top-$k$ candidate KV positions. The corresponding original KV states are then gathered for sparse attention.

This token-wise selection allows Adamas to reduce the number of KV states accessed and processed for each decoding query while preserving high-quality attention candidates. Comprehensive evaluations show that Adamas maintains near-lossless performance within the effective context length of Llama-3.1-8B-Instruct~\citep{hsieh2024ruler}. In efficiency evaluations, Adamas achieves up to $6.5\times$ self-attention speedup and $1.7\times$ end-to-end decoding speedup at a 32K context length. Adamas also remains effective on long-form reasoning tasks, achieving performance comparable to full attention on both AIME 2024~\citep{aime24} and AIME 2025~\citep{aime25} with Qwen3-8B~\citep{qwen3technicalreport} in reasoning mode. Our main contributions are as follows:

\begin{itemize}
    \item We propose \textbf{Adamas}, a token-level sparse attention mechanism that combines Hadamard transform, bucketization, 2-bit compression and lightweight distance-based candidate selection to efficiently identify relevant KV pairs during decoding.
    \item We develop high-performance GPU kernels for Adamas, enabling up to $6.5\times$ self-attention speedup and $1.7\times$ end-to-end decoding speedup over full attention in long-context decoding.
    \item We conduct comprehensive evaluations across long-context understanding, retrieval, language modeling, and long-form reasoning, together with analyses of candidate-selection quality and the key design choices of Adamas.
\end{itemize}
\section{Preliminaries}
\label{sec:preliminaries}

In this section, we briefly review sparse attention and the Hadamard transform to provide background and facilitate a clearer understanding of our proposed method.

\textbf{Sparse attention} is a variant of the self-attention mechanism. The standard attention computation is defined as
\begin{equation}
\label{eq:attention}
    O = \mathrm{Softmax}(\frac{QK^{\top}}{\sqrt{d}})V,
\end{equation}
where $Q$, $K$, $V$, and $O$ denote the query, key, value, and attention output, respectively. 
Since $QK^{\top}$ computes the query--key similarity, achieving high recall requires identifying the most relevant keys for each query. However, evaluating the similarity between all query--key pairs incurs quadratic complexity with respect to the sequence length, making it impractical for long contexts.

Sparse attention reduces this cost by restricting each query to attend to only a selected subset of keys. In general, let $s_i$ denote an estimated relevance score between a query and the $i$-th key, where $i=1,\ldots,N$ and $N$ is the number of tokens. The selected index set is obtained by
\begin{equation}
\label{eq:sparse_topk}
    \mathcal{I} = \operatorname{Topk}(s),
\end{equation}
where $\operatorname{Topk}(\cdot)$ returns the indices of the $k$ most relevant keys according to the scoring function. The corresponding sparse key and value tensors are then gathered as
\begin{equation}
\label{eq:sparse_kv}
    K_\mathcal{I} = K^{\mathrm{all}}[\mathcal{I}], \qquad V_\mathcal{I} = V^{\mathrm{all}}[\mathcal{I}].
\end{equation}
Sparse attention is finally computed only over the selected tokens:
\begin{equation}
\label{eq:sparse_attention}
    O = \mathrm{Softmax}\left(\frac{QK_\mathcal{I}^{\top}}{\sqrt{d}}\right)V_{\mathcal{I}} .
\end{equation}
 
\textbf{Hadamard transform}, also known as the Walsh–Hadamard transform, is an orthogonal linear transform that performs a change of basis using Walsh functions. It corresponds to multiplication by a Hadamard matrix $\mathbf{H}_d \in \mathbb{R}^{d \times d}$, where $d = 2^r$. The Hadamard matrix has entries in $\{+1, -1\}$, and the normalized Hadamard transform is obtained by scaling $\mathbf{H}_d$ by $1/\sqrt{d}$. The smallest normalized Hadamard matrix is given by:
\begin{equation}
\label{eq:hadamard_matrix}
    \mathbf{H}_2 = \frac{1}{\sqrt{2}} 
    \begin{bmatrix}
    1 & 1 \\
    1 & -1
    \end{bmatrix}
\end{equation}
Higher-order matrices are constructed recursively via the Kronecker product $\mathbf{H}_{2^r} = \mathbf{H}_2 \otimes \mathbf{H}_{2^{r-1}}$. This recursive structure enables an efficient Hadamard transform algorithm that computes the matrix–vector product $\mathbf{H}x$ in $\mathcal{O}(d \log_2 d)$ rather than the naive $\mathcal{O}(d^2)$. For dimensions $d$ that are not exact powers of two, the existence of a Hadamard matrix is not guaranteed. In such cases, one may exploit factorizations $d = 2^r m$, where $m$ is the order of a known Hadamard matrix, and apply the Kronecker construction $\mathbf{H}_d = \mathbf{H}_{2^r} \otimes \mathbf{H}_m$, yielding a transform with complexity $\mathcal{O}(d(m+r))$. 
\section{Method}
\label{sec:method}

\begin{figure*}[t]
    \centering
    \includegraphics[width=0.75\linewidth]{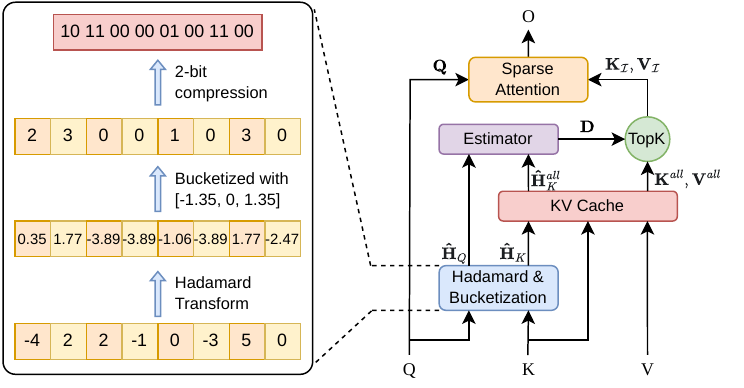}
    \caption{\textbf{Overview of Adamas.} Queries $Q$ and keys $K$ are processed by the Hadamard transform, followed by bucketization and 2-bit compression. The compressed query $\widehat{\mathbf{H}}_Q$ and keys $\widehat{\mathbf{H}}_K$ are compared using a lightweight distance-based estimator to select the top-$k$ candidate positions. The corresponding original keys and values, $K_\mathcal{I}$ and $V_\mathcal{I}$, are then gathered for sparse attention with the original query $Q$. All boxed components are implemented as dedicated GPU kernels for efficiency.}
    \label{fig:Adamas}
\end{figure*}

In this section, we present Adamas, an efficient variant of the Transformer sparse attention mechanism that could substantially reduce computation overhead while preserving model accuracy. We show the workflow of Adamas in Figure~\ref{fig:Adamas}.

\subsection{Design Rationale}

Adamas is designed to construct a compact representation for efficient token-level candidate selection during decoding. We first apply an orthogonal Hadamard transform to queries and keys before discretization. Since $\mathbf{H}\mathbf{H}^{\top} = \mathbf{I}$, applying
the same transform to both preserves their dot products:
\begin{equation}
\label{eq:motivation}
    (Q\mathbf{H})(K\mathbf{H})^{\top} = Q(\mathbf{H}\mathbf{H}^{\top})K^{\top} = QK^{\top}
\end{equation}
Thus, the Hadamard transform changes the representation basis without altering the query--key dot-product similarity before discretization. Directly discretizing query and key vectors, however, can be sensitive to large-magnitude outliers. The Hadamard transform redistributes feature magnitudes more evenly across dimensions and suppresses extreme values, yielding smoother representations that are better suited for coarse quantization~\citep{elhage2023privileged,ashkboos2024quarot}. We therefore bucketize the transformed queries and keys into compact 2-bit codes and use them for lightweight candidate estimation. The structured sign pattern of the Hadamard transform also enables an efficient implementation via the fast Hadamard transform ~\citep{tridao2024fht,ibmmeta2024hadacore}. The effects of the Hadamard transform, compression granularity and candidate selection are evaluated in Sec.~\ref{sec:ablation} and Appendix~\ref{app:candidate_analysis}.

\subsection{Adamas}

As shown in Algorithm~\ref{alg:Adamas}, Adamas augments the standard Transformer attention pipeline with three key innovations: 1) applying the Hadamard transform to queries and keys, yielding what we refer to as Hadamard vectors; 2) bucketization of the Hadamard vectors; and 3) a lightweight distance-based estimation over the compressed representations. These components work jointly to enable faster attention computation with modest memory overhead and near-lossless accuracy. 

\textbf{Hadamard transform applied to queries and keys.}
Formally, given queries and keys $Q, K \in \mathbb{R}^{d}$, and a Hadamard matrix $\mathbf{H} \in \mathbb{R}^{d \times d}$, we compute their Hadamard-transformed representations as
\begin{equation}
\label{eq:hadamard_transform}
    H_Q = Q \mathbf{H}, \qquad H_K = K \mathbf{H},
\end{equation}

\textbf{Bucketization.}
For efficient computation and storage, each element in $H_Q, H_K \in \mathbb{R}^d$ is discretized into one of four levels using predefined thresholds $\{B_1, B_2, B_3\}$, yielding the bucketized Hadamard vectors $\widehat{\mathbf{H}}_Q$ and $\widehat{\mathbf{H}}_K$. The bucketization operator $B(\cdot)$ is defined as
\begin{equation}
    \label{eq:bucketize}
    B(x) = \Sigma_{i=1}^3 \mathbb{I}(x > B_i),
\end{equation}
where $\mathbb{I}(\cdot)$ denotes the indicator function. 

We determine the bucketization scale using a lightweight offline calibration. On a 10\% subsample of representative LongBench datasets, the elements of both $H_Q$ and $H_K$ are approximately zero-centered, with most values concentrated around the origin. We therefore use their empirical standard deviations $\sigma_X, X\in \{ Q,K\}$ as scale parameters and set the three thresholds to $\{-\sigma_X, 0, \sigma_X\}$. The resulting calibrated values are $\sigma_Q = 1.35, \sigma_K=2.26$, which are fixed during inference. Appendix~\ref{app:bucketization} evaluates the sensitivity of this choice.

\begin{algorithm}[!tb]
    \renewcommand{\algorithmicrequire}{\textbf{Input:}}
    \renewcommand{\algorithmicensure}{\textbf{Output:}}
    \renewcommand{\algorithmiccomment}[1]{\hfill$\triangleright$~#1}
    \caption{Workflow of Adamas}
    \label{alg:Adamas}
    \begin{algorithmic}[1]
        \REQUIRE Query $\mathbf{Q}$, Key $\mathbf{K}$, and Value $\mathbf{V} \in \mathbb{R}^{d}$, Hadamard matrix $\mathbf{H} \in \mathbb{R}^{d \times d}$
        \ENSURE Attention output matrix $\mathbf{O}$
        \STATE $\mathbf{H}_Q \gets \mathbf{QH}$; \quad $\mathbf{H}_K \gets \mathbf{KH}$ \COMMENT{Apply Hadamard transform}
        \STATE $\widehat{\mathbf{H}}_Q, \widehat{\mathbf{H}}_K \gets \text{BucketizeCompress}(\mathbf{H}_Q, \mathbf{H}_K)$ \COMMENT{Bucketization and compression}
        \STATE $\widehat{\mathbf{H}}_{K}^{\mathrm{all}}, \mathbf{K}^{\mathrm{all}}, \mathbf{V}^{\mathrm{all}}\gets \text{KVCache.update}(\widehat{\mathbf{H}}_K, \mathbf{K}, \mathbf{V})$
        \STATE $D \gets \text{DistanceEstimator}(\widehat{\mathbf H}_Q,\widehat{\mathbf H}^{\mathrm{all}}_K)$ \COMMENT{Estimate query–key relevance}
        \STATE $I \gets \operatorname{Topk}_{\min}(D)$ \COMMENT{Select top-$k$ candidate indices}
        \STATE $\mathbf{K}_\mathcal{I}, \mathbf{V}_\mathcal{I} \gets \mathbf{K}^{\mathrm{all}}[I], \mathbf{V}^{\mathrm{all}}[I]$ \COMMENT{Gather selected KV states}
        \STATE $\mathbf{O} \gets \text{SparseAttention}(\mathbf{Q}, \mathbf{K}_\mathcal{I}, \mathbf{V}_\mathcal{I})$
        \STATE \textbf{return} $\mathbf{O}$
    \end{algorithmic}
\end{algorithm}

\textbf{Distance-based candidate estimation.} After bucketization, each element of $\widehat{\mathbf{H}}_Q$ and $\widehat{\mathbf{H}}_K$ is represented by a 2-bit code, and every eight codes are packed into a single 16-bit value. The compressed $\widehat{\mathbf{H}}_K$ is stored alongside the original KV cache, incurring only a small additional memory overhead. We estimate the relevance of each key using a lightweight distance function over the compressed representations. In our default implementation, we use the $\ell_1$ distance:
\begin{equation}
\label{eq:manhattan}
    \mathbf{D}(\widehat{\mathbf{H}}_Q,\widehat{\mathbf{H}}_K)
    =
    \left\lVert
    \widehat{\mathbf{H}}_Q-\widehat{\mathbf{H}}_K
    \right\rVert_1 .
\end{equation}
A smaller distance indicates a higher-priority candidate. Since both $\widehat{\mathbf{H}}_Q$ and $\widehat{\mathbf{H}}_K$ are represented by low-bit integers, the distance can be evaluated using lightweight integer operations over the packed representations, avoiding floating-point similarity computation during candidate selection. The compressed representations and distance estimator are used only to construct a candidate set. The corresponding original key and value states are then gathered for sparse attention. We evaluate the quality of the selected candidate sets in Appendix~\ref{app:candidate_analysis} and the effect of distance metric in Sec.~\ref{sec:ablation}.

\textbf{Top-$k$ selection with compressed representations.}
Following the sparse attention formulation in Eq.~\ref{eq:sparse_topk}--\ref{eq:sparse_attention}, our method estimates token relevance using distances in the compressed representation space. Given the compressed query $\widehat{\mathbf{H}}_Q$ and all compressed keys $\widehat{\mathbf{H}}_K^{\mathrm{all}}$, we compute a distance score $D_i$ for each cached token and select the $k$ positions with the smallest distances: 
\begin{equation}
    \mathcal{I}
    =
    \operatorname{TopK}_{\min}(\mathbf{D}).
\end{equation}
The corresponding original key and value states are then gathered as $\mathbf{K}_\mathcal{I}=\mathbf{K}^{\mathrm{all}}[\mathcal{I}]$ and $\mathbf{V}_\mathcal{I}=\mathbf{V}^{\mathrm{all}}[\mathcal{I}]$ for sparse attention.

\textbf{Group-wise selection for grouped-query attention (GQA).}
For GQA models, we exploit the fact that multiple query heads share the same key-value head. Let $\mathcal{G}_g$ denote the set of query heads associated with the $g$-th key-value head. For each query head $q \in \mathcal{G}_g$, let $D_q \in \mathbb{R}^{N}$ be its distance scores to the $N$ tokens. Instead of performing Top-$k$ selection independently for each query head, we compute a group-level distance score by taking the element-wise minimum over all query heads in the same GQA group:
\begin{equation}
    \label{eq:gqa_reduce}
    D^{(g)}_i = \min_{q \in \mathcal{G}_g} D_{q,i},
    \qquad i = 1,\ldots,N .
\end{equation}
We then select the $k$ positions with the smallest group-level distances $I^{(g)} = \operatorname{Topk}_{\min}\big(D^{(g)}\big).$
The selected indices are shared by all query heads in the same GQA group $I_q = I^{(g)},  \forall q \in \mathcal{G}_g .$
This group-wise selection reduces memory accesses and improves computational efficiency, while maintaining comparable retrieval quality.

\textbf{Customized CUDA Kernels}. To fully exploit the compact representation and bit-wise computation enabled by Adamas, we implement a set of specialized high-performance CUDA kernels that optimize each stage of the attention pipeline. Specifically, we design three customized kernels corresponding to fused Hadamard transform + bucketization + compression, fused distance-based candidate estimation + top-$k$ selection, and sparse attention. In our default $\ell_1$ implementation, the distance estimation is efficiently realized with lightweight integer operations over packed 2-bit codes.

\section{Experiments}
\label{sec:experiments}

In this section, we evaluate Adamas in terms of both efficacy and efficiency. We also conduct ablation studies on the Hadamard transform, bucketization, and distance metrics to better understand the contribution of each component.

\begin{figure*}[t]
    \centering
    \includegraphics[width=0.95\linewidth]{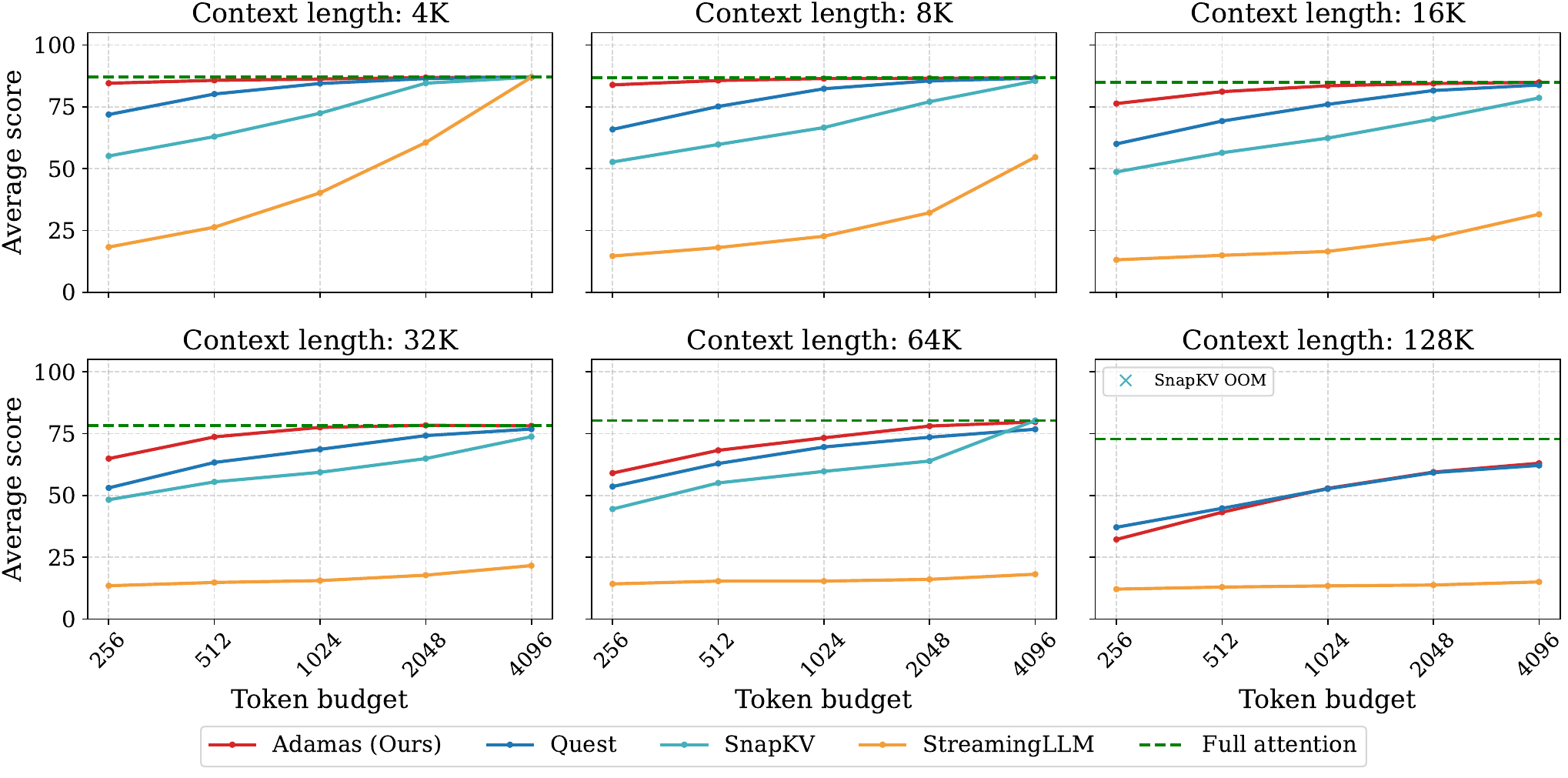}
    \caption{\textbf{Evaluation results on RULER.} Adamas exhibits the smallest performance drop compared to full attention while maintaining high sparsity.}
    \label{fig:ruler}
\end{figure*}

\subsection{Setup}

\textbf{Models.} We use Llama-3.1-8B-Instruct~\citep{llama3_1} and Qwen3-8B~\citep{qwen3technicalreport} as representative GQA models, LongChat-7b-v1.5-32k~\citep{longchat2023} and Yarn-Llama-2-7b-128k~\citep{peng2024yarn} as representative MHA models.

\textbf{Tasks.} 
We consider four categories of evaluation benchmarks. 
First, we evaluate long-context understanding using RULER~\citep{hsieh2024ruler}
and LongBench~\citep{bai2024longbench}. RULER provides synthetic stress tests for long-context
capabilities, whereas LongBench covers diverse real-world tasks: single-document QA (NarrativeQA~\citep{kovcisky2018narrativeqa}, Qasper~\citep{dasigi-etal-2021-dataset}, MultiFieldQA~\citep{bai2024longbench}), multi-document QA (HotpotQA~\citep{yang-etal-2018-hotpotqa}), summarization (GovReport~\citep{huang-etal-2021-efficient}), and few-shot learning (TriviaQA~\citep{joshi-etal-2017-triviaqa}). 
Second, we evaluate long-form mathematical reasoning on AIME 2024 and AIME 2025 using Qwen3-8B in reasoning mode. We use greedy decoding and repeat each evaluation three times, obtaining consistent results across runs. These benchmarks allow us to examine whether sparse attention remains effective over long reasoning trajectories.
Third, we evaluate on the passkey retrieval task~\citep{peng2024yarn}, which measures retrieval capability in long contexts. 
Fourth, we use PG19~\citep{rae2019compressive}, a long-form language modeling dataset, to assess long-context language-modeling quality.

\textbf{Baselines.} For fair comparison, we benchmark against training-free sparse attention methods: StreamingLLM~\citep{xiao2023efficient} for static-sparse-mask method, and SnapKV~\citep{li2024snapkv} and Quest~\citep{tang2024quest} for dynamic-KV-selection methods.

\subsection{Efficacy evaluation}

\textbf{RULER.} We evaluate Llama-3.1-8B-Instruct on all RULER tasks and report the average score. As shown in Figure~\ref{fig:ruler}, Adamas consistently outperforms other baselines across context lengths and token budgets. Within the 32k effective context length of Llama-3.1-8B-Instruct (\citep{hsieh2024ruler}), Adamas achieves near-lossless performance, closely matching full attention scores. In comparison, StreamingLLM consistently underperforms full attention because its KV cache eviction strategy discards many task-relevant tokens. SnapKV shows mediocre accuracy under small and moderate budgets and only becomes competitive at a 4096-token budget, while Quest performs well when the token budget exceeds 1024 but degrades sharply under tighter budgets. These results indicate that Adamas can more reliably identify and preserve task-relevant KV pairs across context lengths.

\begin{table*}[t]
    \centering
    \caption{Performance comparison on LongBench across models with different attention mechanisms (GQA and MHA). Scores denote normalized score (\%) over budgets ranging from 256 to 4096, normalized by full-attention performance. The best results are shown in \textbf{bold}.}
    \begin{tabular}{lcccc}
        \toprule
        Method  & \makecell{Llama-3.1-8B-Instruct \\ (GQA)} & \makecell{Qwen3-8B \\ (GQA)} & \makecell{LongChat-7b-v1.5-32k \\ (MHA)} & Average \\
        \midrule
         Quest & 86.96 & 76.92 & 94.97 & 86.28 \\
         SnapKV & 84.86 & 83.64 & 86.44 & 84.98 \\
         StreamingLLM & 82.32 & 81.22 & 48.61 & 70.72 \\
         \midrule
         Adamas (Ours) & \textbf{97.65} & \textbf{98.04} & \textbf{100.65} & \textbf{98.78}\\
        \bottomrule
    \end{tabular}
    \label{tab:longbench_avg}
\end{table*}

\textbf{LongBench.} Table~\ref{tab:longbench_avg} reports the LongBench results across models with different attention mechanisms, where scores are normalized by full-attention performance. Overall, Adamas achieves the best average performance, reaching 98.78\%, which substantially outperforms Quest (86.28\%), SnapKV (84.98\%), and StreamingLLM (70.72\%). These results show that Adamas preserves nearly full-attention performance not only on GQA models but also on MHA models, demonstrating its robustness across attention architectures. Detailed results are provided in Appendix~\ref{app:longbench}.

\begin{wraptable}{r}{0.5\linewidth}
    \centering
    \caption{
        \textbf{Long-reasoning evaluation on AIME 2024 and AIME 2025
        with Qwen3-8B in reasoning mode.}
        Scores denote the number of correctly solved problems out of 30.
    }
    \label{tab:aime}
    \begin{tabular}{lcccc}
        \toprule
        Benchmark & 1024 & 2048 & 4096 & Full Attn. \\
        \midrule
        AIME 2024 & \textbf{23} & 21 & 20 & 20 \\
        AIME 2025 & 17 & 19 & \textbf{21} & 19 \\
        \bottomrule
    \end{tabular}
\end{wraptable}
\textbf{Long-reasoning evaluations.}
We evaluate Adamas on both AIME 2024 and the more recent AIME 2025 benchmark using Qwen3-8B in reasoning mode, with $\texttt{max\_new\_tokens}=38{,}912$ to accommodate long reasoning traces.
As shown in Table~\ref{tab:aime}, Adamas remains competitive with full attention on both benchmarks. On AIME 2024, it obtains 23/30 with a 1024-token budget and 20/30 with a 4096-token budget, compared with 20/30 under full attention. On AIME 2025, Adamas scores 17/30, 19/30, and 21/30 with budgets of 1024, 2048, and 4096, respectively, compared with 19/30 under full attention. These results show that Adamas preserves long-form reasoning performance even when attention is restricted to a small subset of the KV cache.

\textbf{Passkey results.} 
We evaluate passkey retrieval on Llama-3.1-8B-Instruct, Qwen3-8B and Yarn-Llama-2-7b-128k with 100k-token inputs. 
As shown in Table~\ref{tab:passkey}, Adamas consistently achieves the best or comparable performance across most token budgets, with particularly large gains in low-budget regimes.
Detailed comparisons and analysis are provided in Appendix~\ref{app:passkey}.

\textbf{Perplexity results.} 
We evaluate perplexity on the PG19 dataset using Llama-3.1-8B-Instruct, Qwen3-8B, and LongChat-7b-v1.5-32k with a sequence length of 32K. 
Across all models, Adamas consistently achieves perplexity close to full attention, demonstrating strong accuracy and robustness under long-context settings. 
Detailed results are provided in Appendix~\ref{app:ppl}.

\subsection{Efficiency evaluation}

We conduct efficiency evaluations with Llama-3.1-8B-Instruct (GQA) and LongChat-7b-v1.5-32k (MHA) on NVIDIA RTX A6000 and NVIDIA H800.

\begin{figure*}[t]
    \centering
    \includegraphics[width=\linewidth]{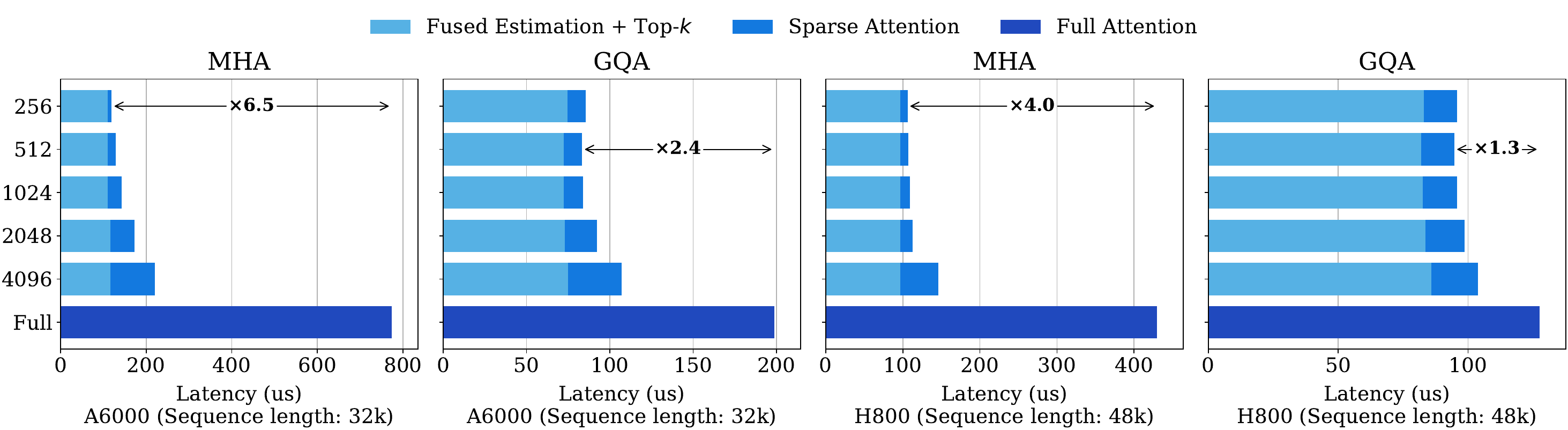}
    \caption{Kernel-level breakdown of Adamas self-attention compared to full attention (FlashInfer).}
    \label{fig:kernels}
\end{figure*}

\textbf{Kernels.} As shown in Figure~\ref{fig:kernels}, Adamas achieves consistent acceleration across different token budgets and sequence lengths. Specifically, Adamas delivers up to $6.5\times$ speedup over full attention with MHA and up to $2.4\times$ speedup with GQA, both implemented using FlashInfer~\citep{ye2025flashinfer}, while maintaining near-lossless accuracy. Since GQA reduces the number of key-value heads, it already provides substantial acceleration compared with MHA; nevertheless, Adamas still achieves additional speedup under GQA, demonstrating its robustness across different attention architectures. The fused Hadamard transform and bucketization-compression introduce negligible overhead and are therefore omitted from further analysis.

\begin{figure}[t]
    \centering
    \includegraphics[width=\linewidth]{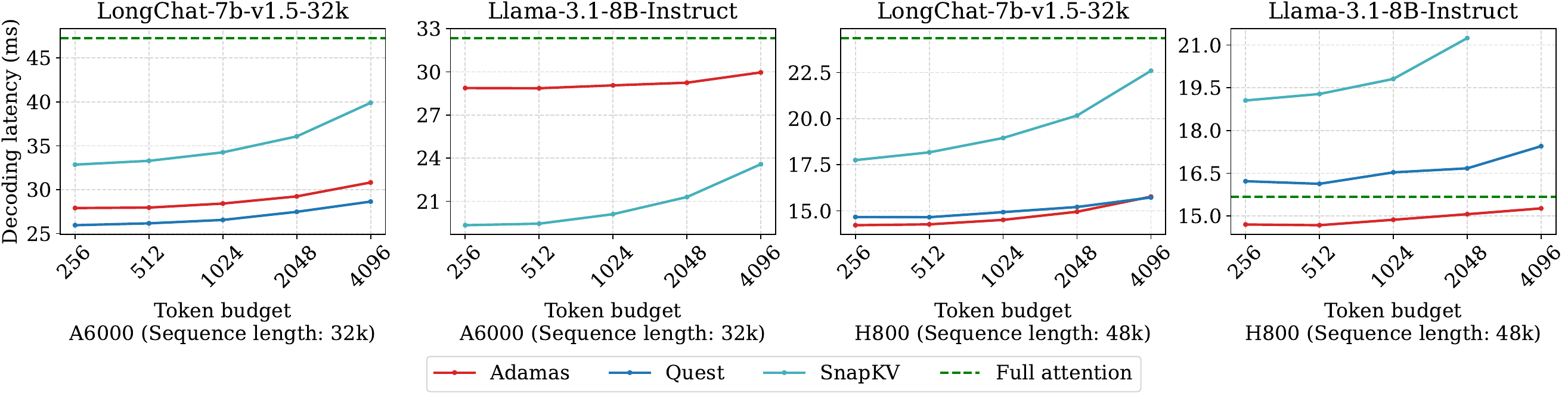}
    \caption{End-to-end decoding latency (ms) comparison. Out-of-Memory (OOM) results omitted.}
    \label{fig:e2e}
\end{figure}

\textbf{End-to-end.}
Figure~\ref{fig:e2e} shows that Adamas provides stable end-to-end decoding acceleration across all token budgets, GPU platforms, and model architectures. As the token budget increases from $256$ to $4096$, Adamas maintains consistently low latency and achieves up to $1.8\times$ speedup on MHA models and $1.1\times$ speedup on GQA models. The gain on GQA is particularly notable: since GQA already reduces the number of key-value heads and therefore substantially lowers the attention cost compared with MHA, the additional speedup achieved by Adamas demonstrates its robustness and practical acceleration capability under more efficient attention architectures.

In comparison, on A6000 with Llama-3.1-8B-Instruct at a 32K context length,
Quest encounters OOM and therefore cannot complete the decoding evaluation. SnapKV provides only limited acceleration in most settings. On H800 with Llama-3.1-8B-Instruct at a 48K context length, both Quest and SnapKV are slower than full attention. For Quest, this is partly due to its lack of native GQA support: its current implementation converts the shared GQA key-value heads into an MHA-style representation before candidate selection and sparse attention, introducing additional memory and computation overhead that becomes non-negligible at long context lengths. In contrast, Adamas performs group-wise candidate selection directly over the native GQA structure without expanding the KV heads to MHA. At 56K, both Quest and SnapKV encounter out-of-memory failures on Llama-3.1-8B-Instruct, whereas Adamas completes the evaluation across all token budgets. Detailed latency results are provided in Appendix~\ref{app:e2e}.

\subsection{Ablation study}
\label{sec:ablation}

\begin{figure*}[t]
    \centering
    \includegraphics[width=\linewidth]{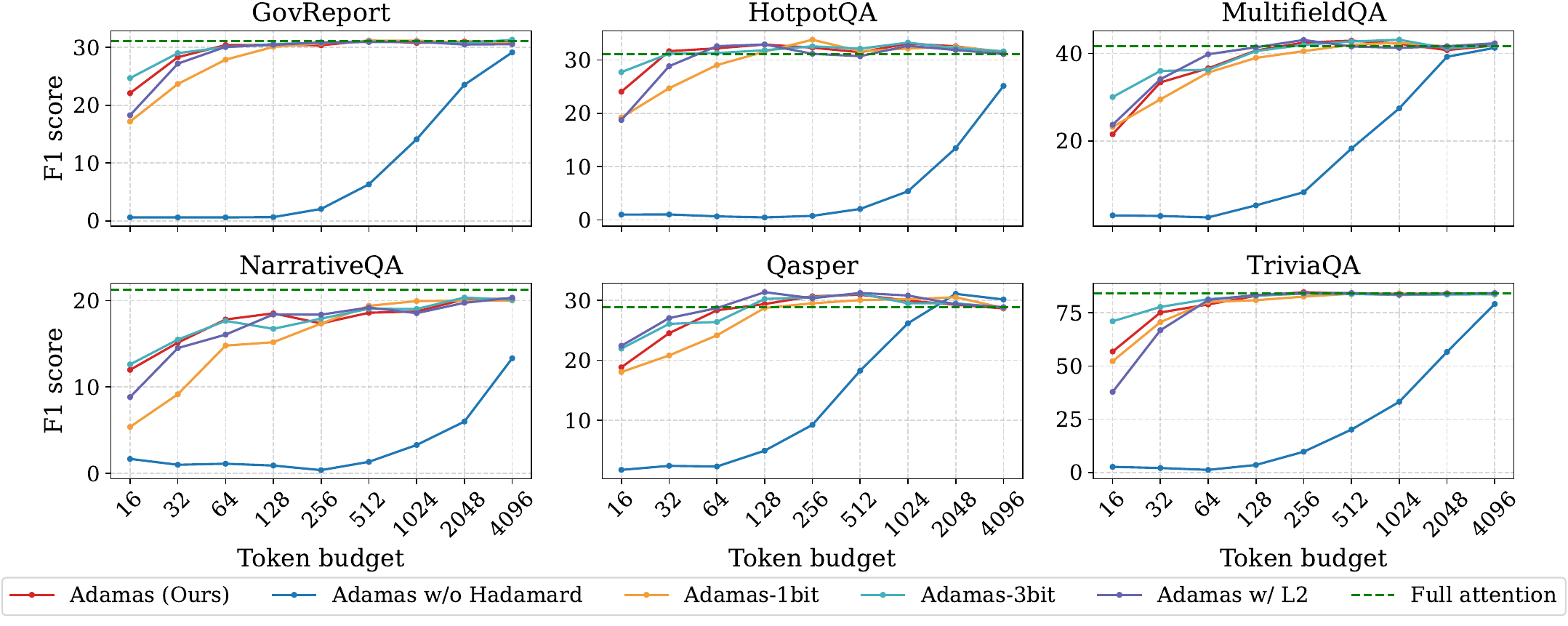}
    \caption{Ablation studies on the Hadamard transform, bucketization, and distance metrics.}
    \label{fig:ablation}
\end{figure*}

In this part, we conduct ablation studies on the Hadamard transform, bucketization, and distance metrics. We benchmark three variants of Adamas on LongBench using LongChat-7b-v1.5-32k (other models show similar results): (1) Adamas without the Hadamard transform, (2) Adamas with different bucketization granularities, and (3) Adamas with $\ell_2$ distance (replacing the default $\ell_1$ distance). The results are shown in Figure~\ref{fig:ablation}, with detailed data provided in Appendix~\ref{app:ablation}.

\textbf{Hadamard transform.} As illustrated in Figure~\ref{fig:ablation}, Adamas without the Hadamard transform (Adamas w/o Hadamard) directly bucketizes queries and keys before distance-based candidate estimation, achieving near-zero scores across multiple datasets under small token budgets. Its performance improves only gradually with larger budgets and remains below the baseline even at a 4096-token budget. This behavior indicates that the model fails to generate accurate answers without the Hadamard transform, underscoring its critical role in Adamas: Hadamard transform suppresses extreme outliers, and thereby mitigates the information loss introduced by bucketization.

\textbf{Bucketization.}  To investigate how bucketization granularity affects accuracy, we compare three variants: 1-bit (Adamas-1bit), 2-bit (Adamas), and 3-bit (Adamas-3bit). As shown in Figure~\ref{fig:ablation}, Adamas and Adamas-3bit achieve nearly identical performance, with Adamas-3bit showing only a slight advantage. Adamas-1bit performs comparably to Adamas when the token budget exceeds $256$, but its accuracy drops significantly under smaller budgets, suggesting that retaining more bits helps preserve information. 
However, the marginal improvement of Adamas-3bit over Adamas highlights diminishing returns when increasing bit width. Since higher bit widths also incur greater storage costs, we adopt 2-bit bucketization as a favorable accuracy–storage trade-off.

\textbf{Distance metrics.} 
We further examine the sensitivity of Adamas to the choice of distance metric by replacing the default $\ell_1$ estimator with $\ell_2$ distance while keeping all other components unchanged. As shown in Figure~\ref{fig:ablation}, the two variants achieve comparable accuracy across datasets and token budgets, with neither metric consistently dominating the other. This result suggests that both metrics provide sufficiently informative signals for candidate selection, and that Adamas is not particularly sensitive to the specific distance function used. We therefore use $\ell_1$ throughout the main experiments because it admits a simple implementation over the packed 2-bit representations.
\section{Related Work}
\label{sec:related_work}

A large amount of work has explored sparsity in attention mechanisms to improve computational efficiency, with a primary focus on how to effectively select critical KV pairs. Existing methods can be broadly categorized into three groups:

\textbf{Static sparse methods.} These methods adopt predefined sparsity patterns, such as sliding windows or attention sinks. For example, infLLM~\citep{xiao2024infllm} relies on the sliding window pattern, while StreamingLLM~\citep{xiao2023efficient} and DUOAttention~\citep{xiao2025duoattention} combine sliding window with attention sink to preserve accuracy. Moreover, DUOAttention requires an extra training phase to determine streaming/retrieval heads, which introduces substantial training overhead. MoA~\citep{fu2025moa} further refines this idea by profiling each attention head across layers to adjust the window span and then incorporating attention sinks to build the sparse mask. Although these static designs yield high throughput, their fixed structures risk discarding tokens that are critical to a given query~\citep{tang2024quest}, often resulting in accuracy degradation under constrained token budgets, particularly on retrieval-intensive tasks.

\textbf{Dynamic KV selection methods.} To better preserve accuracy, these approaches select KV pairs adaptively based on the input query. Quest~\citep{tang2024quest} estimates page importance using an upper-bound approximation of attention weights and selects the top-$k$ pages for sparse attention, though its page-level granularity may miss critical tokens. MInference~\citep{jiang2024minference} defines three dynamic patterns and performs online estimation per attention head to select the proper pattern, while FlexPrefill~\citep{lai2025flexprefill} adaptively determines sparse patterns and selection ratios per head to accelerate long-sequence prefilling. Both methods, however, primarily target the prefilling stage rather than decoding and are therefore complementary to Adamas. We further validate this complementarity by combining Adamas with MInference in Appendix~\ref{app:minference}. SeerAttention~\citep{gao2024seerattention} introduces a learnable gating mechanism that activates important blocks within the attention map, but at the cost of training additional parameters, which complicates deployment by preventing the model from being used out-of-the-box at inference.

\textbf{Training-based sparse methods.} Beyond static and dynamic masking, an alternative approach integrates sparsity into the attention mechanism through specialized training procedures. Reformer~\citep{kitaev2020reformer} replaces standard dot-product attention with locality-sensitive hashing (LSH), reducing complexity from $O(L^2)$ to $O(L\log L)$. NSA~\citep{yuan-etal-2025-native} adopts a dynamic hierarchical sparse strategy that combines coarse-grained token compression, fine-grained token selection, and sliding window via gating to balance global context and local precision. MoBA~\citep{lu2025moba} incorporates a Mixture-of-Experts (MoE) design~\citep{shazeer2017outrageously} into sparse attention, using a gating mechanism to select historical blocks. While these methods often achieve better accuracy and efficiency than standard full attention, they require pretraining the model, which is computationally costly and limits their applicability. In contrast, our method is training-free and can be directly applied to pretrained LLMs, offering broad compatibility without retraining overhead.

\section{Conclusion}
\label{sec:conclusion}

In this paper, we present Adamas, an accurate and efficient token-level sparse attention mechanism for long-context inference. Adamas combines the Hadamard transform, bucketization, 2-bit compression, and lightweight distance-based candidate selection to construct compact KV subsets for each decoding query. The compressed representations are used for efficient candidate estimation, while the corresponding original KV states are gathered for sparse attention. Comprehensive evaluations show that Adamas maintains near-lossless performance within the effective context length of Llama-3.1-8B-Instruct. In efficiency evaluations, Adamas achieves up to $6.5\times$ self-attention speedup and $1.7\times$ end-to-end decoding speedup at a 32K context length. Beyond standard long-context benchmarks, Adamas also remains effective on long-form reasoning tasks. With Qwen3-8B in reasoning mode, it achieves performance comparable to full attention on both AIME 2024 and the more recent AIME 2025 benchmark across practical token budgets. These results demonstrate that Adamas provides a practical approach to accelerating long-context decoding while preserving strong task performance across diverse workloads.

\section*{Acknowledgments}
This work is supported by
Jiangsu Science and Technology Major Project (\#BG2025005), the Collaborative Innovation Center
of Novel Software Technology and Industrialization, and NanJing Kunpeng\&Ascend Center of Cultivation.
Jingwei Xu (\texttt{jingweix@nju.edu.cn}) is the corresponding author.

\section*{AI use statement}

The core idea, methodology, and algorithmic design of Adamas were developed
by the authors. We used generative AI tools to assist with selected
implementation and evaluation tasks, including code modification and
optimization, experiment monitoring and log inspection, and manuscript
editing for clarity and presentation. The tools were also used to help
review the presentation of mathematical arguments and summarize or interpret
experimental results. All AI-assisted code, analyses, and text were reviewed
and verified by the authors. The authors take full responsibility for the
final implementation, experiments, results, and claims presented in this
paper.

\section*{Reproducibility statement}

We release the anonymized source code and evaluation scripts for Adamas at
the repository linked in Section~\ref{sec:experiments}. The release includes
the implementation of the Adamas inference pipeline, benchmark evaluation
scripts, and efficiency-measurement utilities used in our experiments.
Section~\ref{sec:method} describes the algorithm and implementation design,
while Section~\ref{sec:experiments} specifies the evaluated models,
benchmarks, baselines, token budgets, and experimental settings.
The appendix provides additional details on bucketization calibration and
sensitivity, candidate-selection quality, retrieval and perplexity
evaluations, computational and memory-access analysis, detailed LongBench and
latency results, ablation studies, and stress tests. Together, the released
code and documented experimental settings are intended to support
reproduction of the reported accuracy and efficiency results.

\bibliographystyle{plainnat}
\bibliography{references}

@misc{anthropic2025claude4sonnet,
  note={\url{https://www.anthropic.com/news/1m-context}, Accessed: 2025-09-04},
  title={Claude Sonnet 4 now supports 1M tokens of context},
  author={Anthropic},
  year={2025}
}

@misc{openai2025gpt5,
  note={\url{https://openai.com/index/introducing-gpt-5-for-developers/}, Accessed: 2025-09-04},
  title={Introducing GPT‑5 for developers},
  author={OpenAI},
  year={2025}
}

@inproceedings{chang2024booookscore,
    title={BooookScore: A systematic exploration of book-length summarization in the era of {LLM}s},
    author={Yapei Chang and Kyle Lo and Tanya Goyal and Mohit Iyyer},
    booktitle={The Twelfth International Conference on Learning Representations},
    year={2024},
    url={https://openreview.net/forum?id=7Ttk3RzDeu}
}

@inproceedings{nijkamp2023codegen,
    title={CodeGen: An Open Large Language Model for Code with Multi-Turn Program Synthesis},
    author={Erik Nijkamp and Bo Pang and Hiroaki Hayashi and Lifu Tu and Huan Wang and Yingbo Zhou and Silvio Savarese and Caiming Xiong},
    booktitle={The Eleventh International Conference on Learning Representations },
    year={2023},
    url={https://openreview.net/forum?id=iaYcJKpY2B_}
}

@inproceedings{dainese2024generating,
    title={Generating Code World Models with Large Language Models Guided by Monte Carlo Tree Search},
    author={Nicola Dainese and Matteo Merler and Minttu Alakuijala and Pekka Marttinen},
    booktitle={The Thirty-eighth Annual Conference on Neural Information Processing Systems},
    year={2024},
    url={https://openreview.net/forum?id=9SpWvX9ykp}
}

@inproceedings{wu2025mmqa,
    title={{MMQA}: Evaluating {LLM}s with Multi-Table Multi-Hop Complex Questions},
    author={Jian Wu and Linyi Yang and Dongyuan Li and Yuliang Ji and Manabu Okumura and Yue Zhang},
    booktitle={The Thirteenth International Conference on Learning Representations},
    year={2025},
    url={https://openreview.net/forum?id=GGlpykXDCa}
}

@inproceedings{wang-etal-2024-leave,
    title = "Leave No Document Behind: Benchmarking Long-Context {LLM}s with Extended Multi-Doc {QA}",
    author = "Wang, Minzheng  and
      Chen, Longze  and
      Cheng, Fu  and
      Liao, Shengyi  and
      Zhang, Xinghua  and
      Wu, Bingli  and
      Yu, Haiyang  and
      Xu, Nan  and
      Zhang, Lei  and
      Luo, Run  and
      Li, Yunshui  and
      Yang, Min  and
      Huang, Fei  and
      Li, Yongbin",
    editor = "Al-Onaizan, Yaser  and
      Bansal, Mohit  and
      Chen, Yun-Nung",
    booktitle = "Proceedings of the 2024 Conference on Empirical Methods in Natural Language Processing",
    month = nov,
    year = "2024",
    address = "Miami, Florida, USA",
    publisher = "Association for Computational Linguistics",
    url = "https://aclanthology.org/2024.emnlp-main.322/",
    doi = "10.18653/v1/2024.emnlp-main.322",
    pages = "5627--5646"
}

@inproceedings{collabllm2025,
    title={CollabLLM: From Passive Responders to Active Collaborators},
    author={Shirley Wu and Michel Galley and Baolin Peng and Hao Cheng and 
            Gavin Li and Yao Dou and Weixin Cai and James Zou and 
            Jure Leskovec and Jianfeng Gao},
    booktitle={International Conference on Machine Learning (ICML)},
    year={2025}
}

@article{zaheer2020big,
  title={Big bird: Transformers for longer sequences},
  author={Zaheer, Manzil and Guruganesh, Guru and Dubey, Kumar Avinava and Ainslie, Joshua and Alberti, Chris and Ontanon, Santiago and Pham, Philip and Ravula, Anirudh and Wang, Qifan and Yang, Li and others},
  journal={Advances in neural information processing systems},
  volume={33},
  pages={17283--17297},
  year={2020}
}

@article{vaswani2017attention,
  title={Attention is all you need},
  author={Vaswani, Ashish and Shazeer, Noam and Parmar, Niki and Uszkoreit, Jakob and Jones, Llion and Gomez, Aidan N and Kaiser, {\L}ukasz and Polosukhin, Illia},
  journal={Advances in neural information processing systems},
  volume={30},
  year={2017}
}

@article{xiao2023efficient,
  title={Efficient streaming language models with attention sinks},
  author={Xiao, Guangxuan and Tian, Yuandong and Chen, Beidi and Han, Song and Lewis, Mike},
  journal={arXiv preprint arXiv:2309.17453},
  year={2023}
}

@article{xiao2024infllm,
  title={Infllm: Training-free long-context extrapolation for llms with an efficient context memory},
  author={Xiao, Chaojun and Zhang, Pengle and Han, Xu and Xiao, Guangxuan and Lin, Yankai and Zhang, Zhengyan and Liu, Zhiyuan and Sun, Maosong},
  journal={Advances in Neural Information Processing Systems},
  volume={37},
  pages={119638--119661},
  year={2024}
}

@inproceedings{xiao2025duoattention,
    title={DuoAttention: Efficient Long-Context {LLM} Inference with Retrieval and Streaming Heads},
    author={Guangxuan Xiao and Jiaming Tang and Jingwei Zuo and junxian guo and Shang Yang and Haotian Tang and Yao Fu and Song Han},
    booktitle={The Thirteenth International Conference on Learning Representations},
    year={2025},
    url={https://openreview.net/forum?id=cFu7ze7xUm}
}

@inproceedings{fu2025moa,
  title={Mixture of Attention Spans: Optimizing LLM Inference Efficiency with Heterogeneous Sliding-Window Lengths},
  author={Fu, Tianyu and Huang, Haofeng and Ning, Xuefei and Zhang, Genghan and Chen, Boju and Wu, Tianqi and Wang, Hongyi and Huang, Zixiao and Li, Shiyao and Yan, Shengen and others},
  booktitle={Second Conference on Language Modeling},
  year={2025}
}

@article{li2024snapkv,
  title={Snapkv: Llm knows what you are looking for before generation},
  author={Li, Yuhong and Huang, Yingbing and Yang, Bowen and Venkitesh, Bharat and Locatelli, Acyr and Ye, Hanchen and Cai, Tianle and Lewis, Patrick and Chen, Deming},
  journal={Advances in Neural Information Processing Systems},
  volume={37},
  pages={22947--22970},
  year={2024}
}

@inproceedings{tang2024quest,
author = {Tang, Jiaming and Zhao, Yilong and Zhu, Kan and Xiao, Guangxuan and Kasikci, Baris and Han, Song},
title = {QUEST: query-aware sparsity for efficient long-context LLM inference},
year = {2024},
publisher = {JMLR.org},
booktitle = {Proceedings of the 41st International Conference on Machine Learning},
articleno = {1955},
numpages = {11},
location = {Vienna, Austria},
series = {ICML'24}
}

@article{jiang2024minference,
  title={Minference 1.0: Accelerating pre-filling for long-context llms via dynamic sparse attention},
  author={Jiang, Huiqiang and Li, Yucheng and Zhang, Chengruidong and Wu, Qianhui and Luo, Xufang and Ahn, Surin and Han, Zhenhua and Abdi, Amir H and Li, Dongsheng and Lin, Chin-Yew and others},
  journal={Advances in Neural Information Processing Systems},
  volume={37},
  pages={52481--52515},
  year={2024}
}

@inproceedings{lai2025flexprefill,
  title={FlexPrefill: A Context-Aware Sparse Attention Mechanism for Efficient Long-Sequence Inference},
  author={Xunhao Lai and Jianqiao Lu and Yao Luo and Yiyuan Ma and Xun Zhou},
  booktitle={The Thirteenth International Conference on Learning Representations},
  year={2025},
  url={https://openreview.net/forum?id=OfjIlbelrT}
}

@article{gao2024seerattention,
  title={Seerattention: Learning intrinsic sparse attention in your llms},
  author={Gao, Yizhao and Zeng, Zhichen and Du, Dayou and Cao, Shijie and Zhou, Peiyuan and Qi, Jiaxing and Lai, Junjie and So, Hayden Kwok-Hay and Cao, Ting and Yang, Fan and others},
  journal={arXiv preprint arXiv:2410.13276},
  year={2024}
}

@inproceedings{kitaev2020reformer,
    title       = {Reformer: The Efficient Transformer},
    author      = {Nikita Kitaev and Lukasz Kaiser and Anselm Levskaya},
    booktitle   = {International Conference on Learning Representations},
    year        = {2020},
    url         = {https://openreview.net/forum?id=rkgNKkHtvB}
}

@article{lu2025moba,
  title={Moba: Mixture of block attention for long-context llms},
  author={Lu, Enzhe and Jiang, Zhejun and Liu, Jingyuan and Du, Yulun and Jiang, Tao and Hong, Chao and Liu, Shaowei and He, Weiran and Yuan, Enming and Wang, Yuzhi and others},
  journal={arXiv preprint arXiv:2502.13189},
  year={2025}
}

@article{shazeer2017outrageously,
  title={Outrageously large neural networks: The sparsely-gated mixture-of-experts layer},
  author={Shazeer, Noam and Mirhoseini, Azalia and Maziarz, Krzysztof and Davis, Andy and Le, Quoc and Hinton, Geoffrey and Dean, Jeff},
  journal={arXiv preprint arXiv:1701.06538},
  year={2017}
}

@inproceedings{yuan-etal-2025-native,
    title = "Native Sparse Attention: Hardware-Aligned and Natively Trainable Sparse Attention",
    author = "Yuan, Jingyang  and
      Gao, Huazuo  and
      Dai, Damai  and
      Luo, Junyu  and
      Zhao, Liang  and
      Zhang, Zhengyan  and
      Xie, Zhenda  and
      Wei, Yuxing  and
      Wang, Lean  and
      Xiao, Zhiping  and
      Wang, Yuqing  and
      Ruan, Chong  and
      Zhang, Ming  and
      Liang, Wenfeng  and
      Zeng, Wangding",
    editor = "Che, Wanxiang  and
      Nabende, Joyce  and
      Shutova, Ekaterina  and
      Pilehvar, Mohammad Taher",
    booktitle = "Proceedings of the 63rd Annual Meeting of the Association for Computational Linguistics (Volume 1: Long Papers)",
    month = jul,
    year = "2025",
    address = "Vienna, Austria",
    publisher = "Association for Computational Linguistics",
    url = "https://aclanthology.org/2025.acl-long.1126/",
    doi = "10.18653/v1/2025.acl-long.1126",
    pages = "23078--23097",
    ISBN = "979-8-89176-251-0"
}

@article{ashkboos2024quarot,
  title={Quarot: Outlier-free 4-bit inference in rotated llms},
  author={Ashkboos, Saleh and Mohtashami, Amirkeivan and Croci, Maximilian L and Li, Bo and Cameron, Pashmina and Jaggi, Martin and Alistarh, Dan and Hoefler, Torsten and Hensman, James},
  journal={Advances in Neural Information Processing Systems},
  volume={37},
  pages={100213--100240},
  year={2024}
}

@misc{llama3_1,
  author = {Meta},
  title = {meta-llama/Llama-3.1-8B-Instruct},
  year = {2024},
  url = {https://huggingface.co/meta-llama/Llama-3.1-8B-Instruct},
  note = {Accessed: 2026}
}

@misc{qwen3technicalreport,
      title={Qwen3 Technical Report}, 
      author={{Qwen Team}},
      year={2025},
      eprint={2505.09388},
      archivePrefix={arXiv},
      primaryClass={cs.CL},
      url={https://arxiv.org/abs/2505.09388}, 
}

@misc{aime24,
      title={American Invitational Mathematics Examination (AIME) 2024}, 
      author={Zhang, Yifan and Math-AI, Team},
      year={2024},
}

@misc{aime25,
      title={American Invitational Mathematics Examination (AIME) 2025}, 
      author={Zhang, Yifan and Math-AI, Team},
      year={2025},
}

@misc{longchat2023,
    title={How Long Can Open-Source LLMs Truly Promise on Context Length?},
    url={https://lmsys.org/blog/2023-06-29-longchat},
    author={Li, Dacheng and Shao, Rulin and Xie, Anze and Sheng, Ying and Zheng, Lianmin and Gonzalez, Joseph E. and Stoica, Ion and Ma, Xuezhe and Zhang, Hao},
    month={June},
    year={2023}
}

@inproceedings{
      peng2024yarn,
      title={Ya{RN}: Efficient Context Window Extension of Large Language Models},
      author={Bowen Peng and Jeffrey Quesnelle and Honglu Fan and Enrico Shippole},
      booktitle={The Twelfth International Conference on Learning Representations},
      year={2024},
      url={https://openreview.net/forum?id=wHBfxhZu1u}
}

@article{hsieh2024ruler,
  title={RULER: What's the Real Context Size of Your Long-Context Language Models?},
  author={Cheng-Ping Hsieh and Simeng Sun and Samuel Kriman and Shantanu Acharya and Dima Rekesh and Fei Jia and Yang Zhang and Boris Ginsburg},
  year={2024},
  journal={arXiv preprint arXiv:2404.06654},
}

@article{rae2019compressive,
  title={Compressive transformers for long-range sequence modelling},
  author={Rae, Jack W and Potapenko, Anna and Jayakumar, Siddhant M and Lillicrap, Timothy P},
  journal={arXiv preprint arXiv:1911.05507},
  year={2019}
}

@inproceedings{bai2024longbench,
    title = "{L}ong{B}ench: A Bilingual, Multitask Benchmark for Long Context Understanding",
    author = "Bai, Yushi and Lv, Xin  and Zhang, Jiajie  and Lyu, Hongchang  and
      Tang, Jiankai  and Huang, Zhidian  and Du, Zhengxiao  and Liu, Xiao  and Zeng, Aohan  and Hou, Lei  and Dong, Yuxiao  and Tang, Jie  and Li, Juanzi",
    booktitle = "Proceedings of the 62nd Annual Meeting of the Association for Computational Linguistics (Volume 1: Long Papers)",
    month = aug,
    year = "2024",
    address = "Bangkok, Thailand",
    publisher = "Association for Computational Linguistics",
    url = "https://aclanthology.org/2024.acl-long.172",
    doi = "10.18653/v1/2024.acl-long.172",
    pages = "3119--3137",
}

@article{kovcisky2018narrativeqa,
  title={The narrativeqa reading comprehension challenge},
  author={Ko{\v{c}}isk{\`y}, Tom{\'a}{\v{s}} and Schwarz, Jonathan and Blunsom, Phil and Dyer, Chris and Hermann, Karl Moritz and Melis, G{\'a}bor and Grefenstette, Edward},
  journal={Transactions of the Association for Computational Linguistics},
  volume={6},
  pages={317--328},
  year={2018},
  publisher={MIT Press One Rogers Street, Cambridge, MA 02142-1209, USA journals-info~…}
}

@inproceedings{yang-etal-2018-hotpotqa,
    title = "{H}otpot{QA}: A Dataset for Diverse, Explainable Multi-hop Question Answering",
    author = "Yang, Zhilin  and
      Qi, Peng  and
      Zhang, Saizheng  and
      Bengio, Yoshua  and
      Cohen, William  and
      Salakhutdinov, Ruslan  and
      Manning, Christopher D.",
    editor = "Riloff, Ellen  and
      Chiang, David  and
      Hockenmaier, Julia  and
      Tsujii, Jun{'}ichi",
    booktitle = "Proceedings of the 2018 Conference on Empirical Methods in Natural Language Processing",
    month = oct # "-" # nov,
    year = "2018",
    address = "Brussels, Belgium",
    publisher = "Association for Computational Linguistics",
    url = "https://aclanthology.org/D18-1259/",
    doi = "10.18653/v1/D18-1259",
    pages = "2369--2380"
}

@inproceedings{dasigi-etal-2021-dataset,
    title = "A Dataset of Information-Seeking Questions and Answers Anchored in Research Papers",
    author = "Dasigi, Pradeep  and
      Lo, Kyle  and
      Beltagy, Iz  and
      Cohan, Arman  and
      Smith, Noah A.  and
      Gardner, Matt",
    editor = "Toutanova, Kristina  and
      Rumshisky, Anna  and
      Zettlemoyer, Luke  and
      Hakkani-Tur, Dilek  and
      Beltagy, Iz  and
      Bethard, Steven  and
      Cotterell, Ryan  and
      Chakraborty, Tanmoy  and
      Zhou, Yichao",
    booktitle = "Proceedings of the 2021 Conference of the North American Chapter of the Association for Computational Linguistics: Human Language Technologies",
    month = jun,
    year = "2021",
    address = "Online",
    publisher = "Association for Computational Linguistics",
    url = "https://aclanthology.org/2021.naacl-main.365/",
    doi = "10.18653/v1/2021.naacl-main.365",
    pages = "4599--4610"
}

@inproceedings{joshi-etal-2017-triviaqa,
    title = "{T}rivia{QA}: A Large Scale Distantly Supervised Challenge Dataset for Reading Comprehension",
    author = "Joshi, Mandar  and
      Choi, Eunsol  and
      Weld, Daniel  and
      Zettlemoyer, Luke",
    editor = "Barzilay, Regina  and
      Kan, Min-Yen",
    booktitle = "Proceedings of the 55th Annual Meeting of the Association for Computational Linguistics (Volume 1: Long Papers)",
    month = jul,
    year = "2017",
    address = "Vancouver, Canada",
    publisher = "Association for Computational Linguistics",
    url = "https://aclanthology.org/P17-1147/",
    doi = "10.18653/v1/P17-1147",
    pages = "1601--1611"
}

@inproceedings{huang-etal-2021-efficient,
    title = "Efficient Attentions for Long Document Summarization",
    author = "Huang, Luyang  and
      Cao, Shuyang  and
      Parulian, Nikolaus  and
      Ji, Heng  and
      Wang, Lu",
    editor = "Toutanova, Kristina  and
      Rumshisky, Anna  and
      Zettlemoyer, Luke  and
      Hakkani-Tur, Dilek  and
      Beltagy, Iz  and
      Bethard, Steven  and
      Cotterell, Ryan  and
      Chakraborty, Tanmoy  and
      Zhou, Yichao",
    booktitle = "Proceedings of the 2021 Conference of the North American Chapter of the Association for Computational Linguistics: Human Language Technologies",
    month = jun,
    year = "2021",
    address = "Online",
    publisher = "Association for Computational Linguistics",
    url = "https://aclanthology.org/2021.naacl-main.112/",
    doi = "10.18653/v1/2021.naacl-main.112",
    pages = "1419--1436"
}

@article{ye2025flashinfer,
    title = {FlashInfer: Efficient and Customizable Attention Engine for LLM Inference Serving},
    author = {
      Ye, Zihao and
      Chen, Lequn and
      Lai, Ruihang and
      Lin, Wuwei and
      Zhang, Yineng and
      Wang, Stephanie and
      Chen, Tianqi and
      Kasikci, Baris and
      Grover, Vinod and
      Krishnamurthy, Arvind and
      Ceze, Luis
    },
    journal = {arXiv preprint arXiv:2501.01005},
    year = {2025},
    url = {https://arxiv.org/abs/2501.01005}
}

@misc{ibmmeta2024hadacore,
  note={\url{https://pytorch.org/blog/hadacore/}, Accessed: 2025-09-23},
  title={HadaCore: Tensor Core Accelerated Hadamard Transform Kernel},
  author={IBM and Meta},
  year={2024}
}

@misc{tridao2024fht,
  note={\url{https://github.com/Dao-AILab/fast-hadamard-transform}, Accessed: 2025-09-23},
  title={Fast Hadamard Transform in CUDA, with a PyTorch interface},
  author={Tri Dao},
  year={2024}
}

@article{elhage2023privileged,
  title={Privileged bases in the transformer residual stream},
  author={Elhage, Nelson and Lasenby, Robert and Olah, Christopher},
  journal={Transformer Circuits Thread},
  pages={24},
  year={2023}
}

\clearpage
\appendix


\section{Bucketization Calibration and Sensitivity}
\label{app:bucketization}

\begin{figure}[ht]
    \centering
    \includegraphics[width=0.95\linewidth]{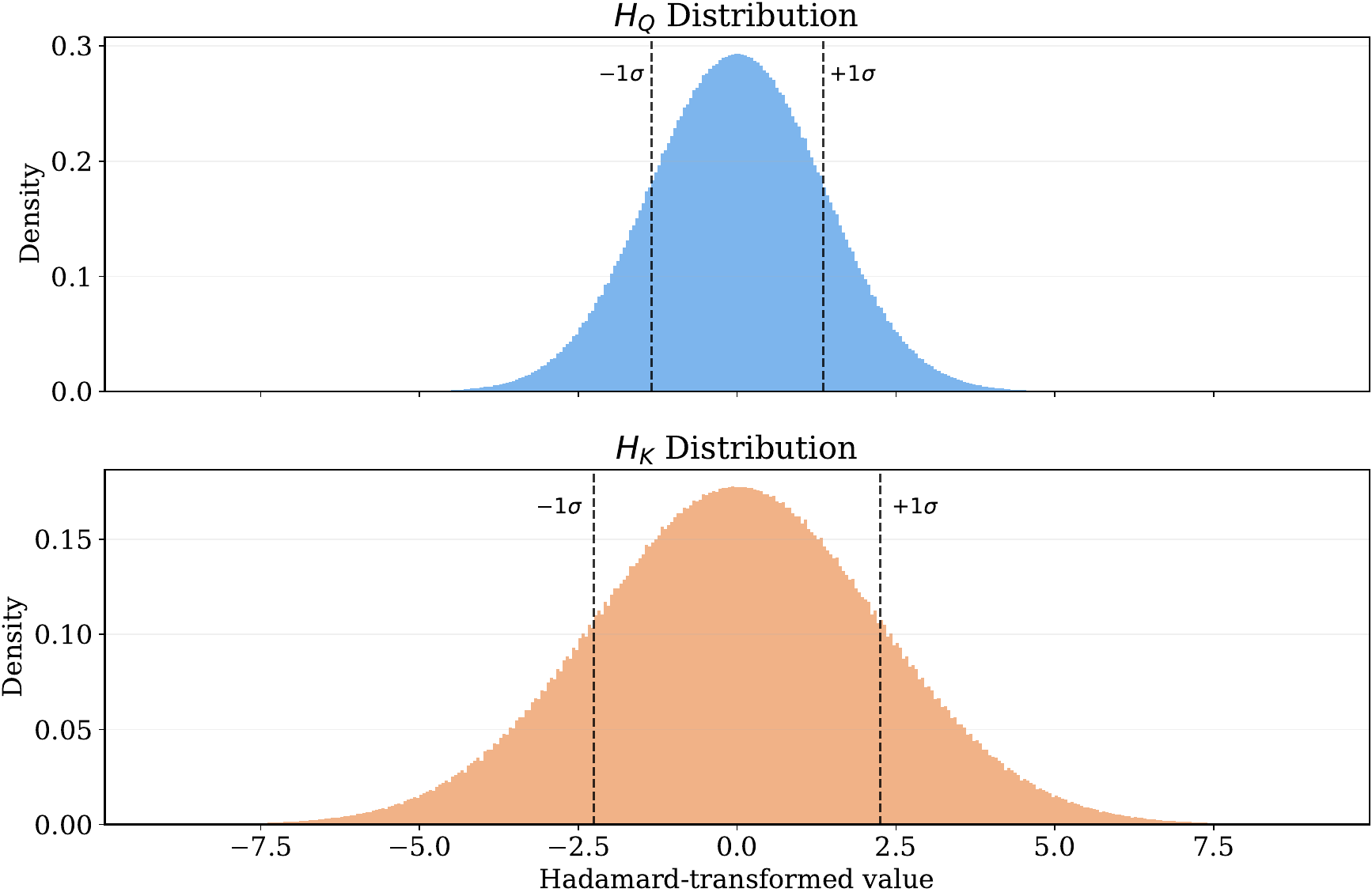}
    \caption{Empirical distributions of the Hadamard-transformed queries $H_Q$ and keys $H_K$ on the calibration subset. Both distributions are approximately zero-centered, with most values concentrated around the origin. The dashed lines indicate $\pm\sigma_X$, which we use as a simple scale-normalized reference for defining the bucketization thresholds.}
    \label{fig:HqHk_distribution}
\end{figure}

In this section, we analyze the calibration and sensitivity of the bucketization thresholds. We first conduct a lightweight offline calibration using a 10\% subsample from each of four representative LongBench datasets, including HotpotQA, TriviaQA, QMSum, and NarrativeQA. Figure~\ref{fig:HqHk_distribution} shows the empirical distributions of the Hadamard-transformed queries and keys. Both are approximately zero-centered, with most values concentrated around the origin.

Rather than tuning the bucket boundaries independently, we use the empirical standard deviation as a simple scale parameter and define the three thresholds as $\{-\sigma_X, 0, \sigma_X\}$ for the Hadamard vector $H_X$. The calibration yields $\sigma_Q=1.35$ and $\sigma_K=2.26$. Once calibrated, these thresholds are fixed during inference and are not adjusted on a per-example or per-query basis.

To evaluate the sensitivity of the bucketization threshold, we vary the threshold multiplier $\alpha \in \{0.5, 1, 1.5, 2\}$, while keeping all other settings fixed. Table~\ref{tab:threshold_sensitivity} reports the results on six LongBench tasks with two model architectures. Performance varies only modestly across this range, and the task-wise optimum does not consistently occur at the same scale. Importantly, $\alpha=1$ achieves the highest macro-average score on both models. These results suggest that Adamas does not rely on a precisely tuned threshold and that $\alpha=1$ provides a simple and robust default.

\begin{table*}[t]
    \centering
    \setlength{\tabcolsep}{4.5pt}
    \renewcommand{\arraystretch}{1.1}
    \caption{\textbf{Sensitivity analysis of $\sigma$-based bucketization thresholds on LongBench.} We vary the threshold multiplier $\alpha \in \{0.5, 1, 1.5, 2\}$ and report results on six LongBench tasks. Best scores for each dataset are in \textbf{bold}. The macro-average results indicate that performance is robust to the threshold scale, with $1\sigma$ providing a strong default across both models.}
    \label{tab:threshold_sensitivity}

    \begin{subtable}[t]{0.48\textwidth}
        \centering
        \caption{Llama-3.1-8B-Instruct}
        \label{tab:threshold_llama}
        \resizebox{\linewidth}{!}{
        \begin{tabular}{lcccc}
            \toprule
            Dataset & $0.5\sigma$ & $1\sigma$ & $1.5\sigma$ & $2\sigma$ \\
            \midrule
            NarrativeQA
                & 27.87 & 27.91 & 28.12 & \textbf{28.31} \\
            Qasper
                & \textbf{42.88} & 42.17 & 41.12 & 42.77 \\
            MultiFieldQA
                & 53.29 & 53.76 & 53.54 & \textbf{54.15} \\
            HotpotQA
                & 53.22 & \textbf{54.97} & 53.15 & 53.54 \\
            GovReport
                & \textbf{33.19} & 33.15 & 32.55 & 32.69 \\
            TriviaQA
                & 90.91 & \textbf{91.37} & 91.33 & 90.61 \\
            \midrule
            \textbf{Macro Avg.}
                & 50.23 & \textbf{50.56} & 49.97 & 50.34 \\
            \bottomrule
        \end{tabular}
        }
    \end{subtable}
    \hfill
    \begin{subtable}[t]{0.48\textwidth}
        \centering
        \caption{LongChat-7b-v1.5-32k}
        \label{tab:threshold_longchat}
        \resizebox{\linewidth}{!}{
        \begin{tabular}{lcccc}
            \toprule
            Dataset & $0.5\sigma$ & $1\sigma$ & $1.5\sigma$ & $2\sigma$ \\
            \midrule
            NarrativeQA
                & 19.13 & \textbf{19.91} & 18.82 & 18.65 \\
            Qasper
                & \textbf{30.40} & 30.05 & 28.83 & 29.67 \\
            MultiFieldQA
                & 41.95 & \textbf{42.45} & 41.13 & 40.32 \\
            HotpotQA
                & 32.24 & \textbf{32.40} & 31.72 & 32.35 \\
            GovReport
                & \textbf{30.91} & 30.71 & 30.51 & 30.76 \\
            TriviaQA
                & 83.87 & \textbf{83.88} & 83.06 & 82.80 \\
            \midrule
            \textbf{Macro Avg.}
                & 39.75 & \textbf{39.90} & 39.01 & 39.09 \\
            \bottomrule
        \end{tabular}
        }
    \end{subtable}
\end{table*}

We additionally repeat the calibration using 20\% and 50\% subsamples from the same datasets and observe nearly unchanged standard deviations, indicating that the calibration statistics can be reliably estimated from a small subset.

\section{Empirical Analysis of Candidate Selection}
\label{app:candidate_analysis}

We further examine the quality of the candidate sets produced by the
distance-based estimator. Since sparse attention ultimately operates on
a selected subset of KV tokens, we measure how much of the attention
information under full attention is retained by the selected candidates.

\paragraph{Evaluation protocol.}
We conduct the analysis on LongChat-7b-v1.5-32k using six LongBench
datasets: Qasper, NarrativeQA, HotpotQA, MultiFieldQA-en, GovReport, and
TriviaQA. We randomly sample 10\% of each dataset, resulting in 115
examples in total. To isolate the quality of the token selector from
differences in generated sequences, all examples are decoded with full
attention. Adamas and Quest~\citep{tang2024quest} are evaluated
off-path on the same queries, keys, and KV cache states produced by
this shared decoding trajectory.

We evaluate token budgets of $\{256, 512, 1024, 2048, 4096\}$. For Quest, we use a page size of 16 tokens. Following the Quest configuration, we exclude the first two layers from the selector analysis for both methods. For Adamas, we use the same calibrated bucketization thresholds as in the main experiments, with $\sigma_Q=1.35$ and $\sigma_K=2.26$. Statistics are collected over all remaining layers, attention heads, and decoding steps.

For a query, let
\begin{equation}
    P_{\mathrm{full}}
    =
    \operatorname{softmax}
    \left(
        \frac{\mathbf{q}\mathbf{K}^{\top}}{\sqrt{d}}
    \right)
\end{equation}
denote the attention distribution under full attention, and let
$\mathcal{S}$ denote the set of tokens selected by a sparse-attention
method. We define the \emph{attention mass recall} as
\begin{equation}
    R_{\mathrm{mass}}(\mathcal{S})
    =
    \sum_{i \in \mathcal{S}}
    P_{\mathrm{full},i}.
    \label{eq:mass_recall}
\end{equation}
This metric directly measures the fraction of full-attention probability
mass retained by the selected candidate set.

We additionally report the recall of high-attention tokens using a
top-$p$ reference set. Specifically, we rank tokens by
$P_{\mathrm{full}}$ and construct the smallest set
$\mathcal{T}_p$ whose cumulative attention mass reaches $p$.
The corresponding set recall is
\begin{equation}
    R_{\mathrm{set}}
    =
    \frac{
        |\mathcal{S}\cap\mathcal{T}_p|
    }{
        |\mathcal{T}_p|
    }.
    \label{eq:set_recall}
\end{equation}
We report $p=0.9$ below.

\begin{table}[t]
    \centering
    \setlength{\tabcolsep}{4.0pt}
    \renewcommand{\arraystretch}{1.1}
    \caption{
        \textbf{Candidate-selection quality on LongBench.}
        Attention mass recall and top-$p$ set recall ($p=0.9$) are
        averaged equally across six datasets. Both Adamas and Quest are
        evaluated on the same full-attention decoding trajectories.
        Higher is better.
    }
    \label{tab:selector_recall}
    \begin{tabular}{c cc c cc}
        \toprule
        & \multicolumn{2}{c}{Mass Recall (\%)} &
        & \multicolumn{2}{c}{Top-$p$ Set Recall (\%)} \\
        \cmidrule(lr){2-3}
        \cmidrule(lr){5-6}
        Budget
        & Adamas & Quest
        &
        & Adamas & Quest \\
        \midrule
        256
        & \textbf{88.97} & 83.88
        &&
        \textbf{76.49} & 69.68 \\

        512
        & \textbf{92.03} & 88.54
        &&
        \textbf{83.80} & 78.59 \\

        1024
        & \textbf{94.70} & 92.27
        &&
        \textbf{90.18} & 86.22 \\

        2048
        & \textbf{96.93} & 95.35
        &&
        \textbf{95.11} & 92.45 \\

        4096
        & \textbf{98.58} & 97.77
        &&
        \textbf{98.18} & 96.84 \\
        \bottomrule
    \end{tabular}
\end{table}

Table~\ref{tab:selector_recall} shows that Adamas consistently retains
more full-attention mass than Quest across all evaluated token budgets.
The difference is most pronounced under high sparsity: with a budget of
256 tokens, Adamas retains $88.97\%$ of the attention mass, compared
with $83.88\%$ for Quest. As the budget increases, both methods approach
full-attention coverage, while Adamas maintains higher recall throughout.

A similar trend is observed for the high-attention token set.
At $p=0.9$, Adamas consistently recalls a larger fraction of the
reference tokens across all budgets. In particular, Adamas exceeds
$90\%$ mass recall with a 512-token budget, whereas Quest first exceeds
this level at 1024 tokens. For top-$p$ set recall, Adamas exceeds
$90\%$ at a budget of 1024, while Quest requires 2048 tokens. At these recall levels, Adamas reaches comparable candidate coverage
with roughly half the token budget required by Quest, highlighting the
benefit of token-level selection under high sparsity.
These results indicate that token-wise distance-based selection can
construct high-quality candidate sets under substantially constrained
token budgets.

\section{Passkey Retrieval Evaluations}
\label{app:passkey}

\textbf{Passkey results.}
We evaluate passkey retrieval on Llama-3.1-8B-Instruct, Qwen3-8B and Yarn-Llama-2-7b-128k with 100k-token inputs. 
As shown in Table~\ref{tab:passkey}, Adamas maintains consistently high accuracy across token budgets and generally outperforms the baselines under constrained budgets.

StreamingLLM performs poorly across all budgets due to its sliding window design, which discards keys outside the retained span. 
Quest achieves strong accuracy with large budgets but degrades sharply under small budgets ($\leq 128$ for 100k), as its page-wise partitioning struggles to recall scattered keys. 
SnapKV also underperforms under limited budgets, indicating instability in highly budget-constrained retrieval.

Overall, these results demonstrate that Adamas can robustly retrieve relevant KV tokens even under severely limited token budgets.

\begin{table*}[t]
    \centering
    \small
    \caption{Passkey retrieval accuracy ($\%$) of Adamas, Quest, SnapKV and StreamingLLM under different token budgets. The best results are in \textbf{bold}. The results highlight the limitations of StreamingLLM’s sliding window design, the budget sensitivity of Quest and SnapKV, and the robustness of Adamas across most budgets. Due to incompatibility between SnapKV and Yarn-Llama-2-7b-128k, we omit the corresponding results.}
    \label{tab:passkey}

    \begin{tabular}{clccccccc}
        \toprule
        Model & Methods & 64 & 128 & 256 & 512 & 1024 & 2048 & 4096 \\
        \midrule
        \multirow{4}{*}{\makecell{Llama-3.1-8B-Instruct \\ (GQA)}} 
        & StreamingLLM  & 1\% & 1\% & 1\% & 1\% & 1\% & 2\% & 4\% \\
        & Quest & \textbf{48\%} & 68\% & 91\% & \textbf{100\%} &\textbf{100\%} & \textbf{100\%} & \textbf{100\%} \\
        & SnapKV & 21\% & 34\% & 49\% & 53\% & 72\% & 86\% & 95\% \\
        \rowcolor{gray!15} & Adamas (Ours) & 45\% & \textbf{91\%} & \textbf{100\%} & \textbf{100\%} &\textbf{100\%} & \textbf{100\%} & \textbf{100\%} \\
        \midrule
        \multirow{4}{*}{\makecell{Qwen3-8B \\ (GQA)}} 
        & StreamingLLM  & 1\% & 1\% & 1\% & 1\% & 1\% & 2\% & 4\% \\
        & Quest & 9\% & 19\% & 76\% & 76\% & 76\% & 86\% & 85\% \\
        & SnapKV & 1\% & 29\% & 69\% & 92\% & 97\% & \textbf{100\%} & \textbf{100\%} \\
        \rowcolor{gray!15} & Adamas (Ours) & \textbf{47\%} & \textbf{91\%} & \textbf{98\%} & \textbf{100\%} & \textbf{100\%} & \textbf{100\%} & \textbf{100\%} \\
        \midrule
        \multirow{4}{*}{\makecell{Yarn-Llama-2-7b-128k \\ (MHA)}}
        & StreamingLLM  & 1\% & 1\% & 1\% & 1\% & 1\% & 2\% & 4\% \\ 
        & Quest  & 25\% & 58\% & 84\% & \textbf{95\%} & \textbf{99\%} & 99\% & 99\% \\
        & SnapKV & \textemdash & \textemdash & \textemdash & \textemdash & \textemdash & \textemdash & \textemdash \\
        \rowcolor{gray!15} & Adamas (Ours) & \textbf{54\%} & \textbf{71\%} & \textbf{87\%} & \textbf{95\%} & 98\% & \textbf{100\%} & \textbf{100\%} \\
        \bottomrule
    \end{tabular}
\end{table*}

\begin{figure}[t]
    \centering
    \includegraphics[width=0.95\linewidth]{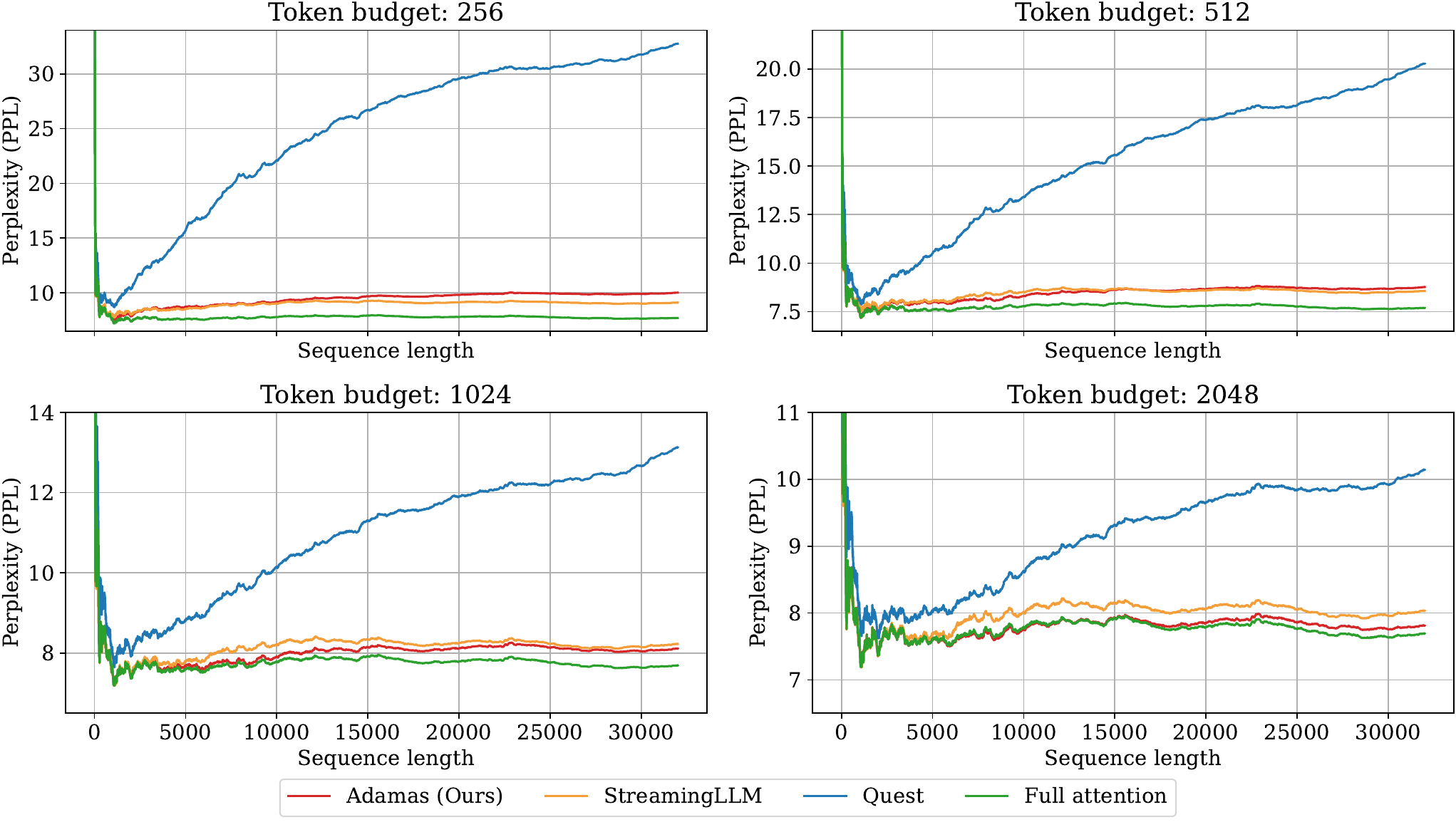}
    \caption{Llama-3.1-8B-Instruct (GQA): perplexity results of StreamingLLM, Quest, Adamas (Ours) and full attention evaluated on PG19 under varying token budgets. Adamas consistently matches full attention and even shows lower perplexity at larger token budget.}
    \label{fig:ppl_llama31}
\end{figure}

\section{Perplexity Evaluations}
\label{app:ppl}

In this section, we further provide results of perplexity evaluation on Llama-3.1-8B-Instruct (GQA), Qwen3-8B (GQA) and LongChat-7b-v1.5-32k (MHA).

\begin{figure}[t]
    \centering
    \includegraphics[width=0.95\linewidth]{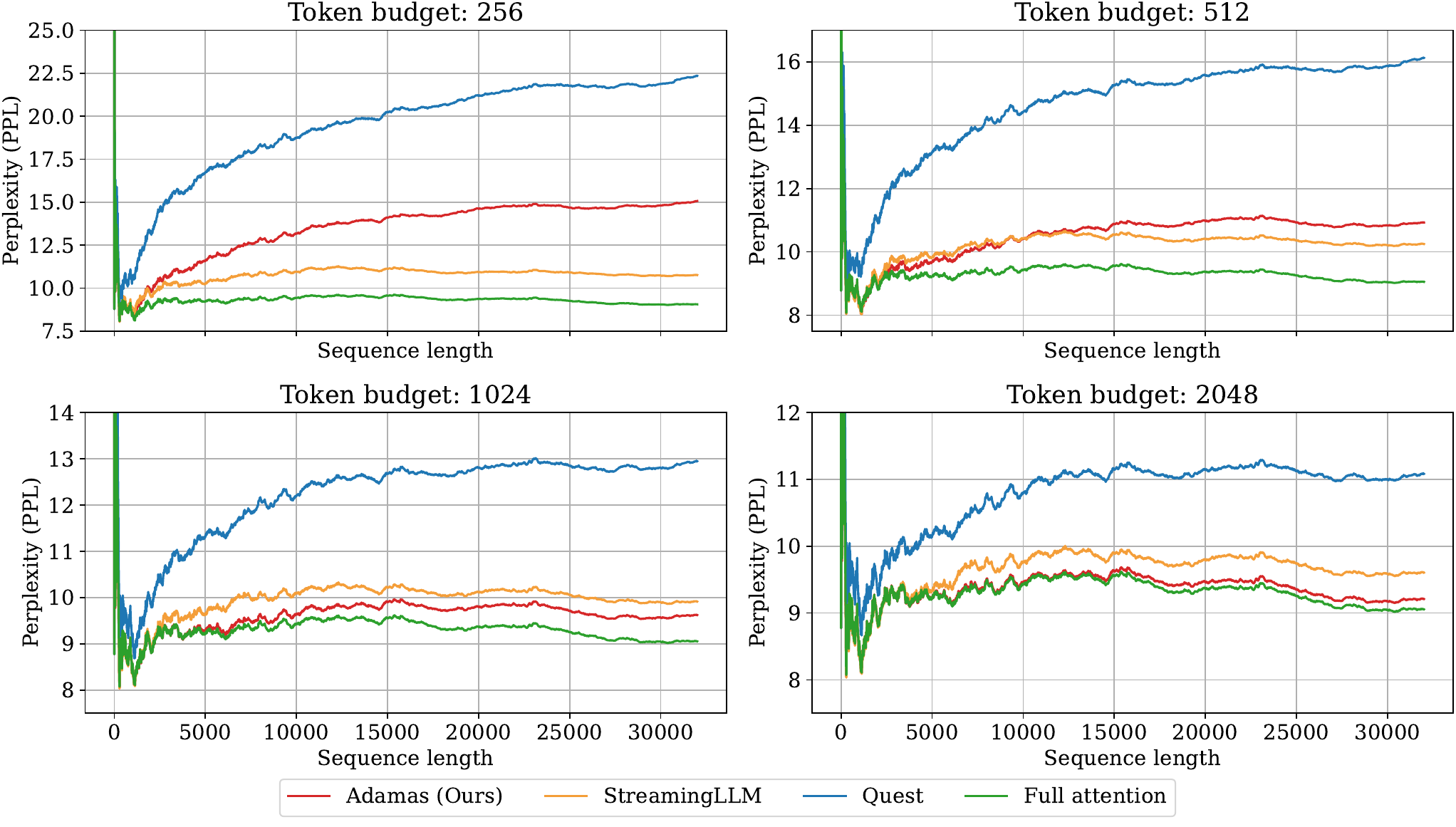}
    \caption{Qwen3-8B (GQA): perplexity results of StreamingLLM, Quest, Adamas (Ours) and full attention evaluated on PG19 under varying token budgets.}
    \label{fig:ppl_qwen3}
\end{figure}

\begin{figure}[t]
    \centering
    \includegraphics[width=0.95\linewidth]{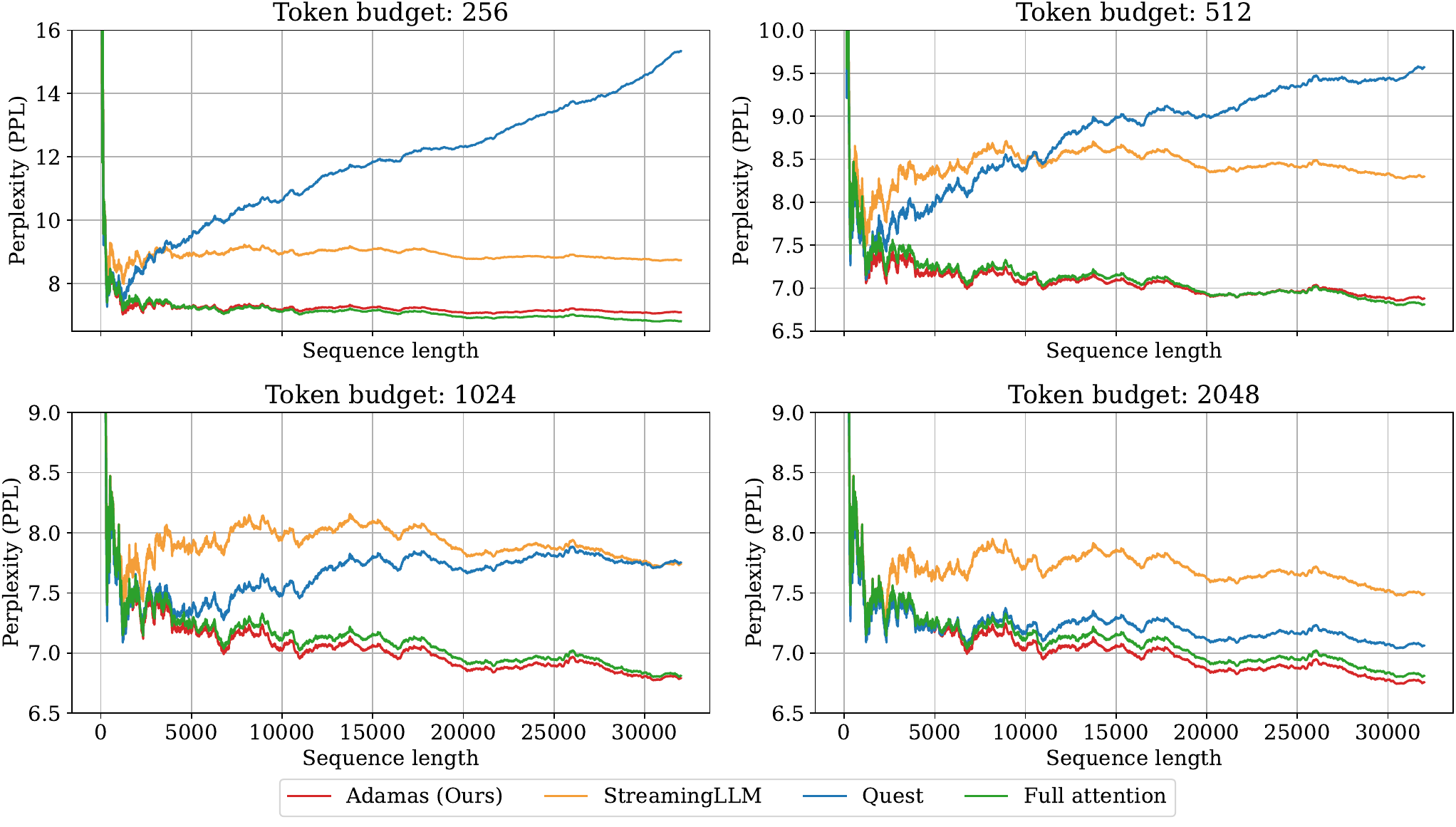}
    \caption{LongChat-7b-v1.5-32k (MHA): perplexity results of StreamingLLM, Quest, Adamas (Ours) and full attention evaluated on PG19 under varying token budgets.}
    \label{fig:ppl_longchat}
\end{figure}

As shown in Figure~\ref{fig:ppl_llama31}, Figure~\ref{fig:ppl_qwen3} and Figure~\ref{fig:ppl_longchat}, Adamas closely tracks full attention across all settings, demonstrating its accuracy and robustness. 
In comparison, Quest yields substantially higher perplexity, as its page-wise strategy introduces many irrelevant KV pairs and can obscure truly relevant ones when the token budget is limited. StreamingLLM maintains stable but consistently higher perplexity due to its eviction strategy, which permanently discards past information. Since SnapKV requires prefilling a long-context input to activate and is therefore incompatible with perplexity evaluation, we omit its result.

\clearpage
\section{Computational Workloads and Memory Access Analysis}
\label{app:analysis}

\subsection{Mathematical notations}

\begin{table}[ht]
  \centering
  \caption{Mathematical notations.}
  \begin{tabular}{rl}
    \toprule
    \textbf{Symbol} & \textbf{Description}.           \\
    \midrule
    $b$       & Batch size                            \\
    $s$       & Sequence length                       \\
    $h$       & Hidden dimension                      \\
    $h_{kv}$  & Hidden dimension of Key and Value     \\
    $d$       & Head dimension                        \\
    $n$       & Number of buckets                     \\
    $P$       & Page size                             \\
    $k$       & Number of selected tokens             \\
    \bottomrule
  \end{tabular}
  \label{tab:notation}
\end{table}

\subsection{Computational workload}

\begin{table}[ht]
    \centering
    \setlength{\tabcolsep}{3.5pt}
    \renewcommand{\arraystretch}{1.1}
    \caption{Computational workload}
    \label{tab:workload}
    \begin{tabular}{llc}
        \toprule
        & Component & FLOPs \\
        \midrule
        \multirow{4}{*}{Full attention}
        & QKV projection & $2bh(h+2h_{kv})$ \\
        & $A = \mathrm{softmax}(QK^{\top})$ & $2bsh$ \\
        & $AV$ & $2bsh$ \\
        & Output projection & $2bh^2$ \\
        \midrule
        \multirow{9}{*}{Quest} 
        & QKV projection & $2bh(h+2h_{kv})$ \\
        & Reduce keys & $2Ph$ \\
        & QK element-wise product & $\frac{2}{P}\cdot bsh$ \\
        & Per channel max & $\frac{1}{P}\cdot bsh$ \\
        & Page sum & $\frac{1}{P}\cdot bsh$ \\
        & Top-$k$ & $\frac{4log_2(\frac{k}{P})}{Pd}\cdot bsh$ \\
        & $A = \mathrm{softmax}(QK^{\top})$ & $2bkh$ \\
        & $AV$ & $2bkh$ \\
        & Output projection & $2bh^2$ \\
        \midrule
        \multirow{8}{*}{Adamas}
        & QKV projection & $2bh(h+2h_{kv})$ \\
        & Hadamard transform& $2bh$ \\
        & Bucketization& $2n\cdot bh$ \\
        & $\ell_1$ distance estimation& $3 bsh$ \\
        & Top-$k$ & $\frac{4log_2(k)}{d}\cdot bsh$ \\
        & $A = \mathrm{softmax}(QK^{\top})$ & $2bkh$ \\
        & $AV$ & $2bkh$ \\
        & Output projection & $2bh^2$ \\
        \bottomrule
    \end{tabular}
\end{table}

Detailed breakdown of computational workload is shown in Table~\ref{tab:workload}. In summary:

\begin{itemize}
\setlength{\parskip}{0.5em}
    \item Computational workload of full attention: $4(h + h_{kv}+s)bh$ FLOPs.
    \item Computational workload of Quest: $[4h + 4h_{kv}+ (\frac{4}{P} + \frac{4log_2(\frac{k}{P})}{Pd})s + 4k]\cdot bh + 2Ph$ FLOPs.
    \item Computational workload of Adamas: $[4h + 4h_{kv} + 2n + 2 + (\frac{4log_2(k)}{d} + 3)s + 4k]\cdot bh$ FLOPs.
\end{itemize}

\subsection{Memory access}

\begin{table*}[ht]
    \centering
    \setlength{\tabcolsep}{3.5pt}
    \renewcommand{\arraystretch}{1.15}
    \caption{Memory access}
    \label{tab:memory}
    \begin{tabular}{llcc}
        \toprule
        & Component & Read access & Write access \\
        \midrule
        \multirow{3}{*}{Full attention}
        & QKV projection & $bh + h(h + 2h_{kv}) + (h + 2h_{kv})$ & $b(h + 2h_{kv})$ \\
        & $A = \mathrm{softmax}(QK^{\top})V$ & $bh + 2bsh_{kv}$ & $bh$ \\
        & Output projection & $bh + h^2$ & $bh$ \\
        \midrule
        \multirow{6}{*}{Quest} 
        & QKV projection & $bh + h(h + 2h_{kv}) + (h + 2h_{kv})$ & $b(h + 2h_{kv})$ \\
        & Reduce keys & $3bh$ & $2Ph$ \\
        & Criticality estimation & $\frac{2bsh}{P} + bh$ & $\frac{bsh}{Pd}$ \\
        & Top-$k$ & $\frac{bsh}{Pd}$ & $\frac{bkh}{Pd}$ \\
        & $A = \mathrm{softmax}(QK^{\top})V$ & $bh + 2bkh_{kv}$ & $bh$ \\
        & Output projection & $bh + h^2$ & $bh$ \\
        \midrule
        \multirow{7}{*}{Adamas}
        & QKV projection & $bh + h(h + 2h_{kv}) + (h + 2h_{kv})$ & $b(h + 2h_{kv})$ \\
        & Hadamard transform& $2bh + 2hd^2$ & $2bh$ \\
        & Bucketization& $2bh$ & $\frac{bh}{4}$ \\
        & $\ell_1$ distance estimation& $\frac{bh+bsh}{8}$ & $\frac{bsh}{d}$ \\
        & Top-$k$ & $\frac{bsh}{d}$ & $\frac{bkh}{d}$ \\
        & $A = \mathrm{softmax}(QK^{\top})V$ & $bh + 2bkh_{kv}$ & $bh$ \\
        & Output projection & $bh + h^2$ & $bh + h^2$ \\
        \bottomrule
    \end{tabular}
\end{table*}

Detailed breakdown of memory access is shown in Table~\ref{tab:memory}. In summary:

\begin{itemize}
\setlength{\parskip}{0.5em}
    \item Memory access of full attention: 
    \begin{itemize}
    \setlength{\parskip}{0.5em}
        \item Read access: $3bh + 2bsh_{kv} + 2h^2 + 2hh_{kv}+h +h_{kv}$
        \item Write access: $3bh + 2bh_{kv}$
    \end{itemize}
    \item Memory access of Quest:
    \begin{itemize}
    \setlength{\parskip}{0.5em}
        \item Read access: $7bh + \frac{2d+1}{Pd}\cdot bsh + 2bkh_{kv} + 2h^2 + 2hh_{kv}+h + 2h_{kv}$
        \item Write access: $(5 + \frac{s+k}{Pd})bh + 2bh_{kv}$
    \end{itemize}
    \item Memory access of Adamas:
    \begin{itemize}
    \setlength{\parskip}{0.5em}
        \item Read access: $\frac{57}{8}bh + \frac{bsh}{d} + 2bkh_{kv} +2d^2 + 2h^2 + 2hh_{kv}+h + 2h_{kv}$
        \item Write access: $(\frac{21}{4} + \frac{s+k}{d})bh + 2bh_{kv}$
    \end{itemize}
\end{itemize}

\clearpage
\section{LongBench Evaluations}
\label{app:longbench}

We present detailed data of Table~\ref{tab:longbench_avg} in Table~\ref{tab:longbench_llama31}, Table~\ref{tab:longbench_qwen3} and Table~\ref{tab:longbench_longchat} for reference. Across the three models, Adamas generally achieves the strongest performance among sparse methods, particularly under limited token budgets.

\begin{table*}[ht]
    \centering
    \setlength{\tabcolsep}{3.5pt}
    \renewcommand{\arraystretch}{1.15}
    \caption{Llama-3.1-8B-Instruct: detailed LongBench evaluation results of StreamingLLM, SnapKV, Quest and Adamas (Ours).  The best results are in \textbf{bold}, and the second-best results are \underline{underlined}. }
    \label{tab:longbench_llama31}
    \begin{tabular}{ccccccccc}
        \toprule
        Datasets & Methods / Budget & 64 & 128 & 256 & 512 & 1024 & 2048 & 4096 \\
        \midrule
        \multirow{5}{*}{\makecell{GovReport \\ (34.51) }}
        &StreamingLLM & \underline{17.20}& 18.65& 20.32& 23.26& 25.20& 28.25& 30.86\\
        &SnapKV & 16.79 & 19.54 & 21.93 & 24.31 & 26.51 & 29.34 & 31.93\\
        &Quest & 13.73& \underline{20.67}& \underline{28.01}& \underline{32.38}& \textbf{33.84}& \underline{34.23}& \textbf{34.90}\\
        &Adamas (Ours) & \textbf{25.01} & \textbf{28.47}& \textbf{30.97}& \textbf{32.84} & \underline{33.29} & \textbf{34.24} & \underline{34.42}\\
        \midrule
        \multirow{5}{*}{\makecell{HotpotQA \\ (55.65)}}
        &StreamingLLM & \textbf{45.78}& \underline{46.84}& \underline{45.79}& 45.66& 47.77& 48.36& 51.28\\
        &SnapKV & 31.77 & 38.34 & 43.72 & \underline{46.04} & 50.77 & \textbf{55.12} & \underline{55.10}\\
        &Quest & 13.33& 26.29& 33.23& 41.93& \underline{51.53}& 51.70& 53.32\\
        &Adamas (Ours) & \underline{43.73}& \textbf{48.84}& \textbf{54.78}& \textbf{54.66} & \textbf{54.66} & \underline{54.81} & \textbf{55.93}\\
        \midrule
        \multirow{5}{*}{\makecell{MultifieldQA \\ (55.22)}}
        &StreamingLLM & \underline{30.58}& 32.27& 32.82& 33.78& 36.47& 38.93& 45.45\\
        &SnapKV & 24.85 & \underline{33.20} & 35.61 & 39.03 & 45.42 & 49.95 & 53.86\\
        &Quest & 15.88& 31.00& \underline{43.19}& \underline{51.32}& \textbf{54.51}& \underline{53.69}& \underline{54.10}\\
        &Adamas (Ours) & \textbf{33.74}& \textbf{46.79}& \textbf{50.17}& \textbf{53.94} & \underline{54.36} & \textbf{55.41} & \textbf{54.94}\\
        \midrule
        \multirow{5}{*}{\makecell{NarrativeQA \\ (29.96) }}
        &StreamingLLM & \textbf{23.16}& \underline{21.68}& \underline{22.67}& \underline{24.48}& \underline{24.40}& 25.30& \underline{27.09}\\
        &SnapKV & 11.18 & 13.15 & 14.34 & 20.07 & 20.10 & 24.76 & 26.81\\
        &Quest & 4.51& 9.52& 14.78& 19.89& 23.43& \underline{26.28}& 26.52\\
        &Adamas (Ours) & \underline{13.20}& \textbf{23.77}& \textbf{24.02}& \textbf{28.20} & \textbf{28.81} & \textbf{28.57} & \textbf{29.93}\\
        \midrule
        \multirow{5}{*}{\makecell{Qasper \\ (43.18)}}
        &StreamingLLM & \underline{20.83}& \underline{20.15}& 22.41& 23.94& 28.42& 34.56& 41.64\\
        &SnapKV & 11.39 & 16.27 & 19.66 & 26.04 & 34.28 & 40.57 & \textbf{42.98}\\
        &Quest & 10.12& 17.15& \underline{31.26}& \underline{35.67}& \underline{39.08}& \underline{42.42}& 41.84\\
        &Adamas (Ours) & \textbf{21.16}& \textbf{32.16}& \textbf{40.26}& \textbf{42.77}& \textbf{42.55} & \textbf{42.69} & \underline{42.59}\\
        \midrule
        \multirow{5}{*}{\makecell{TriviaQA \\ (92.12)}}
        &StreamingLLM & \underline{83.17}& \underline{84.95}& \underline{88.00}& \underline{88.32}& 89.58& \underline{91.81}& 91.78\\
        &SnapKV & 66.86 & 71.23 & 80.04 & 86.67 & \underline{90.09} & 91.19 & \underline{91.88}\\
        &Quest & 21.47& 46.68& 63.42& 74.64& 80.75& 87.47& 91.26\\
        &Adamas (Ours) & \textbf{83.40}& \textbf{89.28}& \textbf{88.44}& \textbf{91.86} & \textbf{92.34} & \textbf{92.27} & \textbf{91.95}\\
        \bottomrule
    \end{tabular}
\end{table*}

\clearpage
\begin{table*}[ht]
    \centering
    \setlength{\tabcolsep}{3.5pt}
    \renewcommand{\arraystretch}{1.15}
    \caption{Qwen3-8B: detailed LongBench evaluation results of StreamingLLM, SnapKV, Quest and Adamas (Ours). The best results are in \textbf{bold}, and the second-best results are \underline{underlined}. }
    \label{tab:longbench_qwen3}
    \begin{tabular}{ccccccccc}
        \toprule
        Datasets & Methods / Budget & 64 & 128 & 256 & 512 & 1024 & 2048 & 4096 \\
        \midrule
        \multirow{5}{*}{\makecell{GovReport \\ (33.38) }}
        &StreamingLLM & \underline{14.27}& 15.62& 18.86& 22.39& 25.81& 28.22& 31.37\\
        &SnapKV & 13.83 & \underline{16.79} & 20.42 & 23.96 & 26.65 & 29.76 & 31.75\\
        &Quest & 8.47 & 13.69 & \underline{21.79}& \underline{28.60}& \underline{32.24}& \underline{32.75}& \underline{32.81}\\
        &Adamas (Ours) & \textbf{18.88}& \textbf{29.38}& \textbf{35.07}& \textbf{34.43} & \textbf{33.98} & \textbf{33.57} & \textbf{33.69}\\
        \midrule
        \multirow{5}{*}{\makecell{HotpotQA \\ (53.65)}}
        &StreamingLLM & \textbf{36.72}& \underline{37.77}& \underline{38.00}& 38.97& 40.60& 44.17& 49.09\\
        &SnapKV & \underline{27.37} & \textbf{37.90} & 36.91 & \underline{41.84} & \underline{47.79} & \underline{50.28} & \textbf{54.26}\\
        &Quest & 4.69& 8.09& 23.35& 32.77& 44.52& 48.39& 50.91\\
        &Adamas (Ours) & 14.90& \textbf{37.90}& \textbf{46.12}& \textbf{56.42} & \textbf{55.22} & \textbf{56.45} & \underline{53.10}\\
        \midrule
        \multirow{5}{*}{\makecell{MultifieldQA \\ (53.09)}}
        &StreamingLLM & \underline{24.45}& \underline{29.20}& 29.23& 30.57& 33.87& 37.55& 43.05\\
        &SnapKV & 21.22 & 27.96 & \underline{33.96} & 38.29 & 43.85 & 46.49 & 49.99\\
        &Quest & 11.85& 22.24& 33.86& \underline{41.18}& \underline{45.98}& \underline{48.48}& \underline{51.25}\\
        &Adamas (Ours) & \textbf{39.22}& \textbf{46.30}& \textbf{50.49}& \textbf{50.18} & \textbf{51.86} & \textbf{52.02} & \textbf{52.55}\\
        \midrule
        \multirow{5}{*}{\makecell{NarrativeQA \\ (24.83) }}
        &StreamingLLM & \textbf{16.78}& \underline{16.71}& \underline{17.42}& \underline{18.01}& \underline{19.95}& \underline{20.92}& 22.18\\
        &SnapKV & \underline{8.50} & 9.30 & 12.64 & 14.86 & 16.96 & 19.78 & \underline{22.68}\\
        &Quest & 1.75& 4.03& 8.07& 13.48& 18.16& 19.59& 20.49\\
        &Adamas (Ours) & 5.67& \textbf{17.15}& \textbf{20.49}& \textbf{22.12} & \textbf{24.22} & \textbf{23.14} & \textbf{24.88}\\
        \midrule
        \multirow{5}{*}{\makecell{Qasper \\ (44.72)}}
        &StreamingLLM & \textbf{23.63}& \underline{24.84}& \underline{27.19}& 29.22& 31.48& 36.99& 41.65\\
        &SnapKV & 10.86 & 16.39 & 18.31 & 24.06 & 31.61 & 38.90 & \underline{44.00}\\
        &Quest & 6.62& 15.28& 24.43& \underline{30.07}& \underline{36.36}& \underline{39.91}& 42.21\\
        &Adamas (Ours) & \underline{21.42}& \textbf{32.20}& \textbf{38.49}& \textbf{42.32}& \textbf{44.16} & \textbf{45.13} & \textbf{44.87}\\
        \midrule
        \multirow{5}{*}{\makecell{TriviaQA \\ (88.59)}}
        &StreamingLLM & \textbf{75.22}& \textbf{79.77}& \underline{83.60}& \underline{86.37}& \underline{87.71}& 88.12& \textbf{88.71}\\
        &SnapKV & \underline{29.52} & \underline{63.39} & 78.99 & 84.30 & 86.89 & \textbf{88.54} & \underline{88.54}\\
        &Quest & 8.95& 18.57& 36.50& 54.53& 68.55& 79.24& 86.57\\
        &Adamas (Ours) & 16.9& 59.55& \textbf{83.80}& \textbf{87.34} & \textbf{88.95} & \underline{88.46} & \underline{88.54}\\
        \bottomrule
    \end{tabular}
\end{table*}

\clearpage
\begin{table*}[ht]
    \centering
    \setlength{\tabcolsep}{3.5pt}
    \renewcommand{\arraystretch}{1.15}
    \caption{LongChat-7b-v1.5-32k: detailed LongBench evaluation results of StreamingLLM, SnapKV, Quest and Adamas (Ours). The best results are in \textbf{bold}, and the second-best results are \underline{underlined}. }
    \label{tab:longbench_longchat}
    \begin{tabular}{ccccccccc}
        \toprule
        Datasets & Methods / Budget & 64 & 128 & 256 & 512 & 1024 & 2048 & 4096 \\
        \midrule
        \multirow{5}{*}{\makecell{GovReport \\ (31.08) }}
        &StreamingLLM & 1.24& 1.28& 1.18& 1.21& 6.18& 24.80& 27.77\\
        &SnapKV & 1.55 & 1.21 & 3.66 & 21.11 & 23.62 & 26.31 & 28.47\\
        &Quest & \underline{3.30}& \underline{11.93}& \underline{22.84}& \underline{27.27}& \underline{29.89}& \textbf{31.19}& \textbf{31.23}\\
        &Adamas (Ours) & \textbf{30.41}& \textbf{30.44}& \textbf{30.32}& \textbf{30.25} & \textbf{31.03} & \underline{31.10} & \underline{30.86}\\
        \midrule
        \multirow{5}{*}{\makecell{HotpotQA \\ (31.01)}}
        &StreamingLLM & \underline{10.72}& 12.02& 13.90& 10.20& 14.56& 24.80& 27.04\\
        &SnapKV & 10.55 & 12.16 & 20.28 & \textbf{35.65} & \textbf{33.90} & \textbf{35.23} & \textbf{33.94}\\
        &Quest & 5.31& \underline{15.49}& \underline{21.54}& 26.64& 30.29& 32.21& \underline{32.93}\\
        &Adamas (Ours) & \textbf{31.18}& \textbf{31.89}& \textbf{31.39}& \underline{33.76} & \underline{32.45} & \underline{32.82} & 31.58\\
        \midrule
        \multirow{5}{*}{\makecell{MultifieldQA \\ (41.96)}}
        &StreamingLLM & 4.14& 5.88& 4.81& 3.95& 11.30& 28.78& 33.63\\
        &SnapKV & \underline{7.53} & 6.24 & 16.74 & \underline{40.98} & 37.76 & 40.03 & \underline{42.82}\\
        &Quest & 6.99& \underline{20.87}& \underline{31.12}& 39.80& \underline{42.03}& \textbf{44.09}& \textbf{43.25}\\
        &Adamas (Ours) & \textbf{36.60}& \textbf{40.67}& \textbf{41.05}& \textbf{44.10} & \textbf{42.76} & \underline{41.98} & 42.38\\
        \midrule
        \multirow{5}{*}{\makecell{NarrativeQA \\ (21.53) }}
        &StreamingLLM & 3.67& 3.47& 4.91& 2.69& 5.45& 16.48& 19.09\\
        &SnapKV & \underline{4.62} & 2.45 & 9.50 & \underline{17.97} & \underline{19.47} & 19.31 & \textbf{20.32}\\
        &Quest & 1.86& \underline{5.98}& \underline{14.2}& 16.28& 17.91& \underline{19.66}& 19.88\\
        &Adamas (Ours) & \textbf{17.81}& \textbf{18.52}& \textbf{18.98}& \textbf{19.82} & \textbf{19.52} & \textbf{20.71} & \underline{20.53}\\
        \midrule
        \multirow{5}{*}{\makecell{Qasper \\ (28.74)}}
        &StreamingLLM & 7.51& 8.03& 7.43& 6.73& 9.56& 21.90& 28.41\\
        &SnapKV & \underline{11.11} & 8.97 & 9.02 & 25.25 & 26.40 & 28.92 & \underline{29.25}\\
        &Quest & 8.67& \underline{18.54}& \underline{26.72}& \textbf{30.89}& \textbf{31.01}& \textbf{31.86}& \textbf{29.79}\\
        &Adamas (Ours) & \textbf{28.33}& \textbf{29.38}& \textbf{31.30}& \underline{30.49} & \underline{30.39} & \underline{29.18} & 28.89\\
        \midrule
        \multirow{5}{*}{\makecell{TriviaQA \\ (83.55)}}
        &StreamingLLM & 2.65& 3.09& 3.84& 9.86& 41.94& 82.42& 83.28\\
        &SnapKV & 0.67 & 9.19 & 55.92 & \textbf{84.01} & 80.88 & 80.12 & 81.25\\
        &Quest & \underline{12.29}& \underline{44.68}& \underline{68.92}& 81.32& \textbf{83.95}& \textbf{85.84}& \textbf{84.94}\\
        &Adamas (Ours) & \textbf{78.91}& \textbf{82.95}& \textbf{83.65}& \underline{83.80} & \underline{83.80} & \underline{84.06} & \underline{84.11}\\
        \bottomrule
    \end{tabular}
\end{table*}

\clearpage
\section{End-to-end Efficiency Evaluation}
\label{app:e2e}

In this section, we present detailed results of Figure~\ref{fig:e2e}.

On A6000, Quest is slightly faster than Adamas on LongChat-32k, while Adamas still achieves approximately $1.5\times$--$1.7\times$ speedup over full attention. On Llama-3.1-32k, SnapKV achieves the lowest latency, whereas Quest is not applicable.

On H800, Adamas exhibits more stable performance across settings. For LongChat-48k and LongChat-56k, Adamas is either the fastest or close to the fastest among the evaluated methods, achieving approximately $1.6\times$--$1.8\times$ speedup over full attention, while SnapKV generally incurs higher latency. For Llama-3.1, Adamas is the only method with complete results at 56k context length, whereas both Quest and SnapKV have missing results. This indicates that Adamas provides better usability and robustness under longer-context settings.

\begin{table}[ht]
\centering
\small
\caption{Decoding latency (ms) measured on A6000. ``--'' denotes OOM failures.}
\label{tab:e2e-latency-a6000}
\begin{tabular}{ccccccccc}
\toprule
\makecell{Context \\ length} & Model & Method & 256 & 512 & 1024 & 2048 & 4096 & Full \\
\midrule
\multirow{3}{*}{32k} & \multirow{3}{*}{LongChat-7b-v1.5-32k} & Quest & 25.95 & 26.17 & 26.55 & 27.48 & 28.65 & \multirow{3}{*}{47.28} \\
& & SnapKV & 32.85 & 33.29 & 34.25 & 36.07 & 39.90 & \\
& & Adamas (Ours) & 27.91 & 27.96 & 28.42 & 29.24 & 30.82 & \\
\midrule
\multirow{3}{*}{32k} & \multirow{3}{*}{Llama-3.1-8B-Instruct} & Quest & -- & -- & -- & -- & -- & \multirow{3}{*}{32.35} \\
& & SnapKV & 19.33 & 19.44 & 20.10 & 21.29 & 23.57 & \\
& & Adamas (Ours) & 28.88 & 28.86 & 29.06 & 29.25 & 29.96 & \\
\bottomrule
\end{tabular}
\end{table}

\begin{table}[ht]
\small
\centering
\caption{Decoding latency (ms) measured on H800. ``--'' denotes OOM failures.}
\label{tab:e2e-latency-h800}
\begin{tabular}{ccccccccc}
\toprule
\makecell{Context \\ length} & Model & Method & 256 & 512 & 1024 & 2048 & 4096 & Full \\
\midrule
\multirow{3}{*}{48k} & \multirow{3}{*}{LongChat-7b-v1.5-32k} & Quest & 14.67 & 14.66 & 14.93 & 15.21 & 15.73 & \multirow{3}{*}{24.37} \\
& & SnapKV & 17.75 & 18.17 & 18.95 & 20.17 & 22.60 & \\
& & Adamas (Ours) & 14.23 & 14.27 & 14.51 & 14.95 & 15.77 & \\
\midrule
\multirow{3}{*}{48k} & \multirow{3}{*}{Llama-3.1-8B-Instruct} & Quest & 16.22 & 16.13 & 16.53 & 16.67 & 17.45 & \multirow{3}{*}{15.68} \\
& & SnapKV & 19.06 & 19.28 & 19.81 & 21.25 & -- & \\
& & Adamas (Ours) & 14.70 & 14.68 & 14.87 & 15.06 & 15.27 & \\
\midrule
\multirow{3}{*}{56k} & \multirow{3}{*}{LongChat-7b-v1.5-32k} & Quest & 15.32 & 15.28 & 15.47 & 15.90 & 16.44 & \multirow{3}{*}{26.79} \\
& & SnapKV & 18.05 & 18.25 & 18.98 & 20.03 & 22.59 & \\
& & Adamas (Ours) & 14.70 & 14.91 & 15.21 & 15.66 & 16.33 & \\
\midrule
\multirow{3}{*}{56k} & \multirow{3}{*}{Llama-3.1-8B-Instruct} & Quest & -- & -- & -- & -- & -- & \multirow{3}{*}{16.49} \\
& & SnapKV & -- & -- & -- & -- & -- & \\
& & Adamas (Ours) & 15.19 & 15.27 & 15.26 & 15.53 & 15.98 & \\
\bottomrule
\end{tabular}
\end{table}

\section{Combining Adamas with MInference}
\label{app:minference}

Adamas primarily accelerates the autoregressive decoding stage, while
existing sparse-attention methods for prefilling optimize a different
phase of inference. To examine whether these optimizations can be
combined, we integrate Adamas with MInference~\citep{jiang2024minference},
using MInference during prefilling and Adamas during decoding. This
configuration requires no additional training or architectural
modification: MInference is enabled for prompt-side attention
computation, and the model switches to Adamas for token-wise KV
selection during autoregressive decoding.

We evaluate the combined pipeline at context lengths of 100K and 128K
across token budgets from 256 to 4096. We additionally include the full
attention setting for reference. Table~\ref{tab:minference_summary}
summarizes the results averaged across token budgets, and Table~\ref{tab:minference_detail} shows detailed results.

\begin{table}[ht]
    \centering
    \setlength{\tabcolsep}{4.5pt}
    \renewcommand{\arraystretch}{1.1}
    \caption{
        \textbf{Latency of Adamas combined with MInference.}
        MInference is applied during prefilling, while Adamas is used
        during decoding. Results are averaged across all evaluated settings. The relative change is measured against Adamas without
        MInference.
    }
    \label{tab:minference_summary}
    \begin{tabular}{lccc|ccc}
        \toprule
        & \multicolumn{3}{c|}{Prefill (s)}
        & \multicolumn{3}{c}{Decode (s/token)} \\
        \cmidrule(lr){2-4}
        \cmidrule(lr){5-7}
        Setting
        & Adamas
        & +MInference
        & Change
        & Adamas
        & +MInference
        & Change \\
        \midrule
        100K
        & 16.656
        & 11.895
        & $-28.6\%$
        & 0.016692
        & 0.016768
        & $+0.45\%$ \\

        128K
        & 25.794
        & 17.269
        & $-33.0\%$
        & 0.018559
        & 0.018661
        & $+0.56\%$ \\

        \midrule
        Average
        & 21.225
        & 14.582
        & $\mathbf{-30.8\%}$
        & 0.017625
        & 0.017714
        & $+0.51\%$ \\
        \bottomrule
    \end{tabular}
\end{table}

As shown in Table~\ref{tab:minference_summary}, combining MInference with Adamas substantially reduces prefill latency
while leaving decoding latency nearly unchanged. Averaged across all
evaluated settings, MInference reduces prefill latency from 21.225 to
14.582, corresponding to a 30.8\% reduction, or approximately a
$1.46\times$ speedup. The improvement becomes slightly larger at longer
contexts, reaching 33.0\% at 128K compared with 28.6\% at 100K.

In contrast, the decoding latency changes by only 0.51\% on average.
This is expected because MInference operates only during the prefill
stage, while decoding continues to use the same Adamas candidate-selection
and sparse-attention pipeline. These results show that the two methods
provide complementary acceleration across different inference stages:
MInference reduces prompt-processing cost, whereas Adamas reduces the
repeated attention cost during autoregressive decoding.

\begin{table*}[ht]
    \centering
    \caption{
        \textbf{Detailed latency (s for prfill and s/token for decode) results for combining Adamas with
        MInference.}
        MInference accelerates the prefill stage, while the decoding
        stage continues to use Adamas. $\Delta$ reports the relative
        latency change after enabling MInference.
    }
    \label{tab:minference_detail}
    \begin{tabular}{ccccc|ccc}
        \toprule
        Token
        & Context
        & Adamas
        & +MInference
        & $\Delta$ Prefill
        & Adamas
        & +MInference
        & $\Delta$ Decode \\
        Budget
        & Length
        & Prefill
        & Prefill
        &
        & Decode
        & Decode
        & \\
        \midrule

        256  & 100K & 16.544 & 11.755 & $-28.9\%$
             & 0.016001 & 0.016073 & $+0.45\%$ \\
        256  & 128K & 25.769 & 17.316 & $-32.8\%$
             & 0.017860 & 0.017976 & $+0.65\%$ \\

        512  & 100K & 16.654 & 11.888 & $-28.6\%$
             & 0.016053 & 0.016153 & $+0.62\%$ \\
        512  & 128K & 25.783 & 17.297 & $-32.9\%$
             & 0.017784 & 0.017936 & $+0.86\%$ \\

        1024 & 100K & 16.687 & 11.890 & $-28.7\%$
             & 0.016207 & 0.016331 & $+0.76\%$ \\
        1024 & 128K & 25.771 & 17.229 & $-33.1\%$
             & 0.018091 & 0.018224 & $+0.74\%$ \\

        2048 & 100K & 16.669 & 11.945 & $-28.3\%$
             & 0.016609 & 0.016546 & $-0.38\%$ \\
        2048 & 128K & 25.777 & 17.275 & $-33.0\%$
             & 0.018322 & 0.018530 & $+1.14\%$ \\

        4096 & 100K & 16.670 & 11.921 & $-28.5\%$
             & 0.017038 & 0.017128 & $+0.53\%$ \\
        4096 & 128K & 25.760 & 17.301 & $-32.8\%$
             & 0.018947 & 0.018891 & $-0.29\%$ \\

        Full & 100K & 16.710 & 11.970 & $-28.4\%$
             & 0.018244 & 0.018377 & $+0.73\%$ \\
        Full & 128K & 25.903 & 17.197 & $-33.6\%$
             & 0.020350 & 0.020407 & $+0.28\%$ \\

        \bottomrule
    \end{tabular}
\end{table*}

As shown in Table~\ref{tab:minference_detail}, the reduction in prefill
latency is consistent across all token budgets, ranging from 28.3\%--28.9\%
at 100K and 32.8\%--33.6\% at 128K. Meanwhile, the decoding differences
remain within approximately $\pm1\%$ for nearly all settings, confirming
that enabling MInference does not materially affect the Adamas decoding
pipeline.

Overall, Adamas can be naturally combined with prefill-oriented sparse
attention methods such as MInference, extending its decoding-time
acceleration to a more complete long-context inference pipeline without
additional training.

\clearpage
\section{Ablation Studies}
\label{app:ablation}

Below we present detailed ablation results evaluated on LongBench with LongChat-7b-v1.5-32k in Table~\ref{tab:ablation}.

\begin{table}[ht]
    \centering
    \setlength{\tabcolsep}{3.5pt}
    \small
    \caption{Ablation results evaluated on LongBench with LongChat-7b-v1.5-32k.}
    \label{tab:ablation}
    \begin{tabular}{lcccccccccc}
        \toprule
        Datasets & Methods / Budget & 16 & 32 & 64 & 128 & 256 & 512 & 1024 & 2048 & 4096 \\
        \midrule
        \multirow{5}{*}{GovReport}&Adamas (Ours) & 22.08& 28.31& \textbf{30.41}& 30.44& 30.37& \textbf{31.18} & 30.77 & \textbf{31.00} & 31.07\\
        &Adamas w/o Hadamard & 0.60& 0.59& 0.59& 0.64& 2.06& 6.32& 14.12& 23.53& 29.14\\
        &Adamas-1bit & 17.17& 23.67& 27.88& 30.09& 30.72& 31.16& \textbf{31.38}& 30.84& 30.70\\
        (31.12)&Adamas-3bit & \textbf{24.69}& \textbf{29.01}& 30.05& \textbf{30.59}& 30.83& 31.01& 30.95& 30.71& \textbf{31.33}\\
        &Adamas w/ L2 distance  & 18.27& 27.19& 30.09& 30.45& \textbf{30.86}& 30.93& 30.96& 30.50& 30.56\\
        \midrule
        \multirow{5}{*}{HotpotQA}&Adamas (Ours) & 24.05& \textbf{31.65}& 32.18& 32.89& 32.30& 31.45 & 33.06 & \textbf{32.56} & 31.41\\
        &Adamas w/o Hadamard & 1.01& 1.04& 0.68& 0.48& 0.77& 2.07& 5.38& 13.47& 25.13\\
        &Adamas-1bit & 19.22& 24.70& 29.04& 31.55& \textbf{33.79}& 31.63& 32.14& 32.47& 31.58\\
        (31.07)&Adamas-3bit &\textbf{27.72}& 31.17& 31.29& 31.82& 32.56& \textbf{32.12}& \textbf{33.25}& 32.21& \textbf{31.61}\\
        &Adamas w/ L2 distance  & 18.73& 28.83& \textbf{32.58}& \textbf{32.91}& 31.15& 30.70& 32.66& 31.86& 31.10\\
        \midrule
        \multirow{5}{*}{MultifieldQA}&Adamas (Ours) & 21.53& 33.38& 36.60& 40.67& 42.49& \textbf{42.93} & 42.22 & 40.77 & 41.83\\
        &Adamas w/o Hadamard & 2.97& 2.84& 2.51& 5.27& 8.28& 18.26& 27.45& 39.24& 41.29\\
        &Adamas-1bit & 23.21& 29.51& 35.64& 38.98& 40.53& 41.99& 42.43& 41.62& 42.01\\
        (41.64)&Adamas-3bit &\textbf{30.04}& \textbf{36.00}& 36.28& 40.54& 41.96& 42.73& \textbf{43.14}& 41.17& 41.86\\
        &Adamas w/ L2 distance  & 23.70& 34.11& \textbf{39.81}& \textbf{41.41}& \textbf{43.08}& 41.60& 41.28& \textbf{41.65}& \textbf{42.33}\\
        \midrule
        \multirow{5}{*}{NarrativeQA}&Adamas (Ours) & 11.98& 15.13& \textbf{17.81}& \textbf{18.52}& 17.36& 18.59 & 18.74 & 20.14 & 20.22\\
        &Adamas w/o Hadamard & 1.68& 1.00& 1.12& 0.91& 0.38& 1.34& 3.29& 6.00& 13.32\\
        &Adamas-1bit & 5.38& 9.16& 14.79& 15.18& 17.34& \textbf{19.39}& \textbf{19.94}& 20.00& 20.01\\
        (21.23)&Adamas-3bit & \textbf{12.61}& \textbf{15.48}& 17.65& 16.73& 17.91& 19.05& 19.03& \textbf{20.35}& 20.09\\
        &Adamas w/ L2 distance  & 8.83& 14.50& 16.06& 18.39& \textbf{18.37}& 19.20& 18.54& 19.74& \textbf{20.33}\\
        \midrule
        \multirow{5}{*}{Qasper}&Adamas (Ours) & 18.83& 24.52& 28.33& 29.38& \textbf{30.68}& 30.93 & 29.94 & 29.26 & 28.65\\
        &Adamas w/o Hadamard & 1.73& 2.40& 2.29& 4.93& 9.24& 18.27& 26.17& \textbf{31.08}& \textbf{30.14}\\
        &Adamas-1bit & 18.04& 20.81& 24.16& 28.73& 29.50& 30.04& 30.18& 30.52& 28.73\\
        (28.89)&Adamas-3bit & 21.95& 26.07& 26.39& 30.23& 30.52& 31.12& 29.50& 29.51& 28.81\\
        &Adamas w/ L2 distance  & \textbf{22.41}& \textbf{27.04}& \textbf{28.70}& \textbf{31.36}& 30.35& \textbf{31.23}& \textbf{30.80}& 29.36& 28.86\\
        \midrule
        \multirow{5}{*}{TriviaQA}&Adamas (Ours) & 56.77& 75.13& 78.91& 82.95& \textbf{84.67}& 83.99 & 83.36 & 83.75 & 83.95\\
        &Adamas w/o Hadamard & 2.63& 2.08& 1.21& 3.52& 9.73& 20.14& 33.15& 56.62& 79.13\\
        &Adamas-1bit & 52.26& 70.62& 80.08& 80.88& 82.59& 83.90& \textbf{84.11}& \textbf{84.22}& 83.63\\
        (84.25)&Adamas-3bit & \textbf{70.99}& \textbf{77.74}& \textbf{81.40}& 82.95& 83.99& 83.79& 83.46& 83.55& 83.68\\
        &Adamas w/ L2 distance  &37.82 &66.80 & 80.97 &\textbf{83.35} &84.21 &\textbf{84.29} &83.60 &84.03 &\textbf{84.25} \\
        \bottomrule
    \end{tabular}
\end{table}

\section{Needle-in-a-Haystack Stress Tests}
\label{app:NIAH}

\begin{figure}[ht]
    \centering

    \begin{subfigure}[t]{0.48\linewidth}
        \centering
        \includegraphics[width=\linewidth]{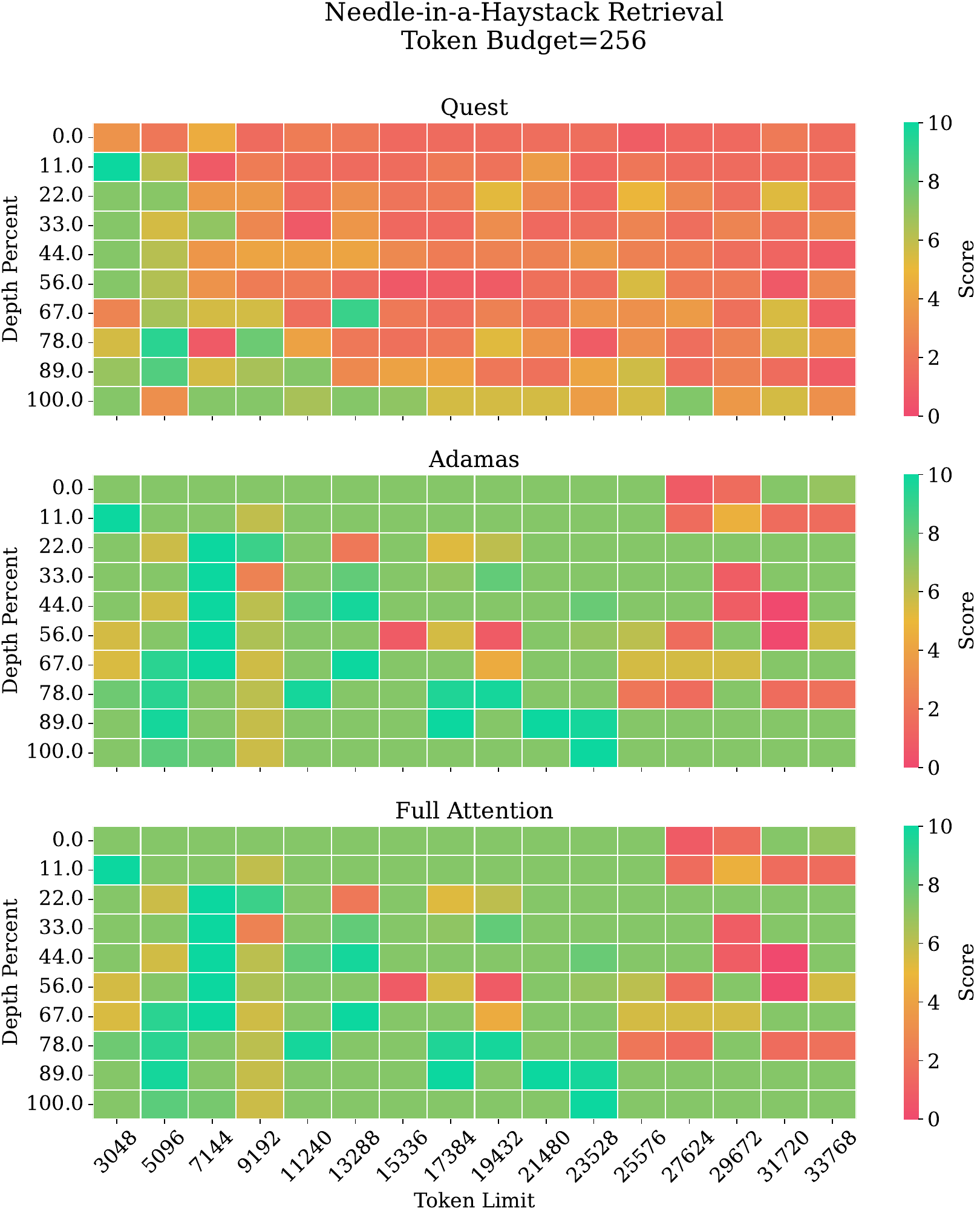}
    \end{subfigure}
    \hfill
    \begin{subfigure}[t]{0.48\linewidth}
        \centering
        \includegraphics[width=\linewidth]{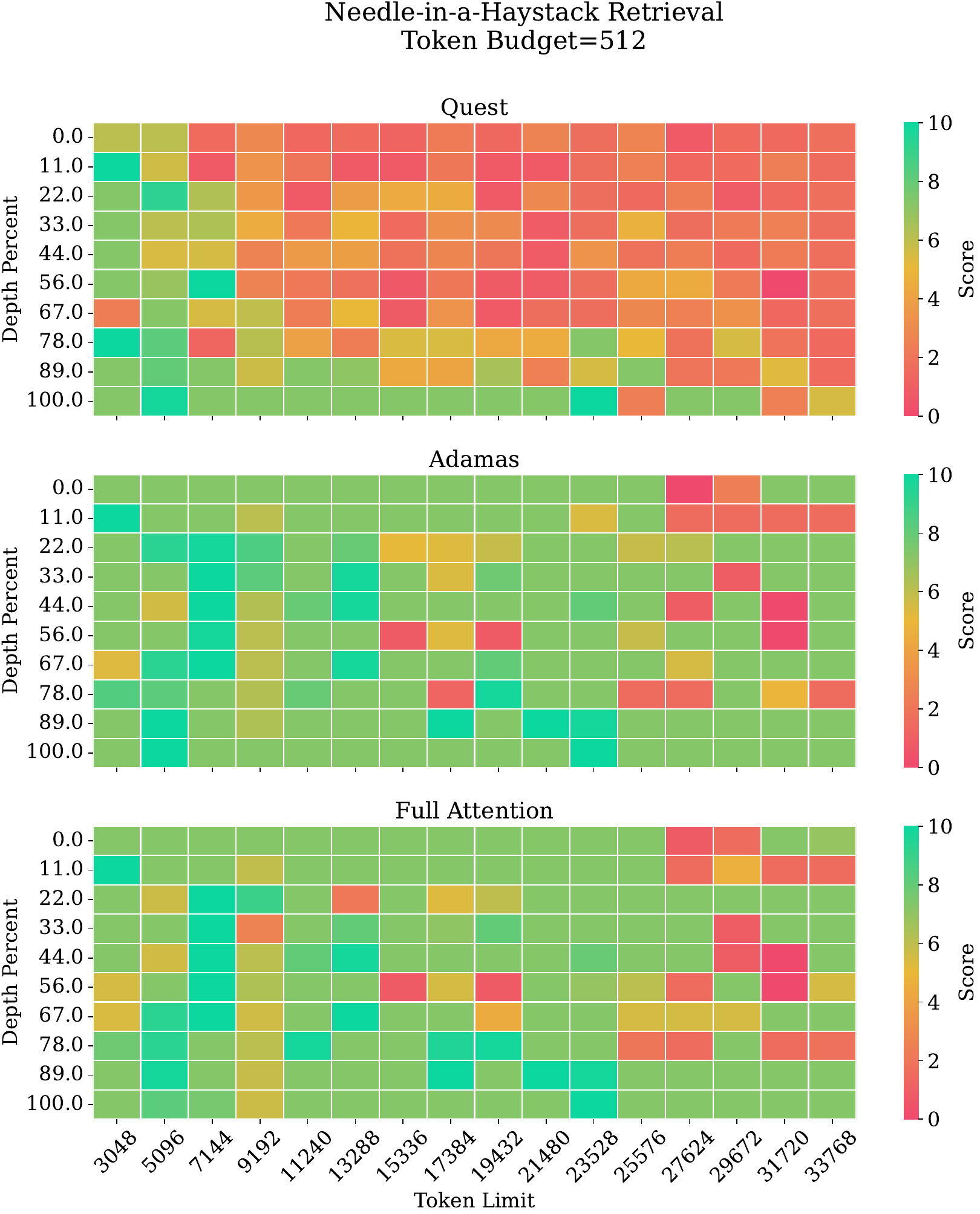}
    \end{subfigure}

    \vspace{0.5em}

    \begin{subfigure}[t]{0.48\linewidth}
        \centering
        \includegraphics[width=\linewidth]{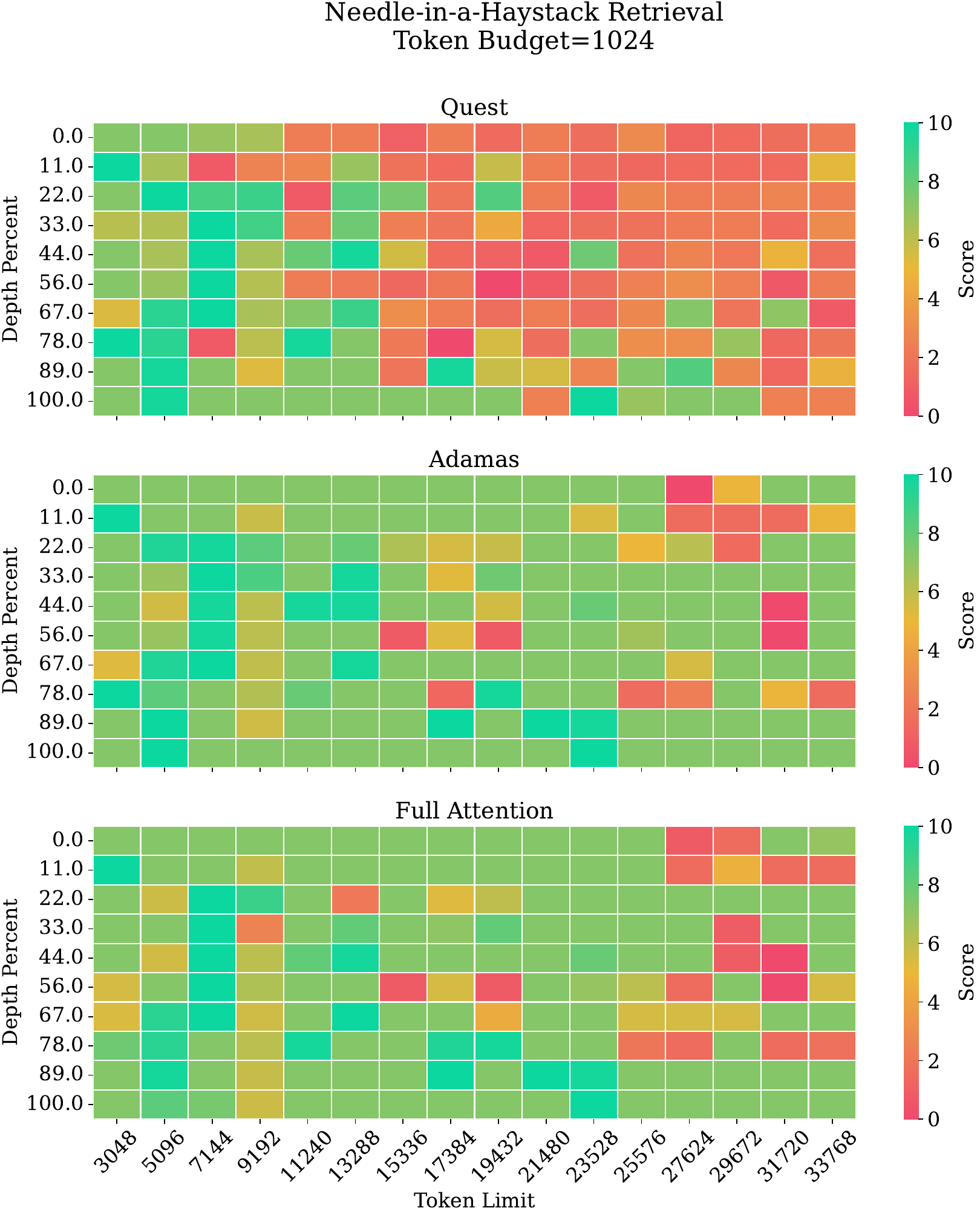}
    \end{subfigure}
    \hfill
    \begin{subfigure}[t]{0.48\linewidth}
        \centering
        \includegraphics[width=\linewidth]{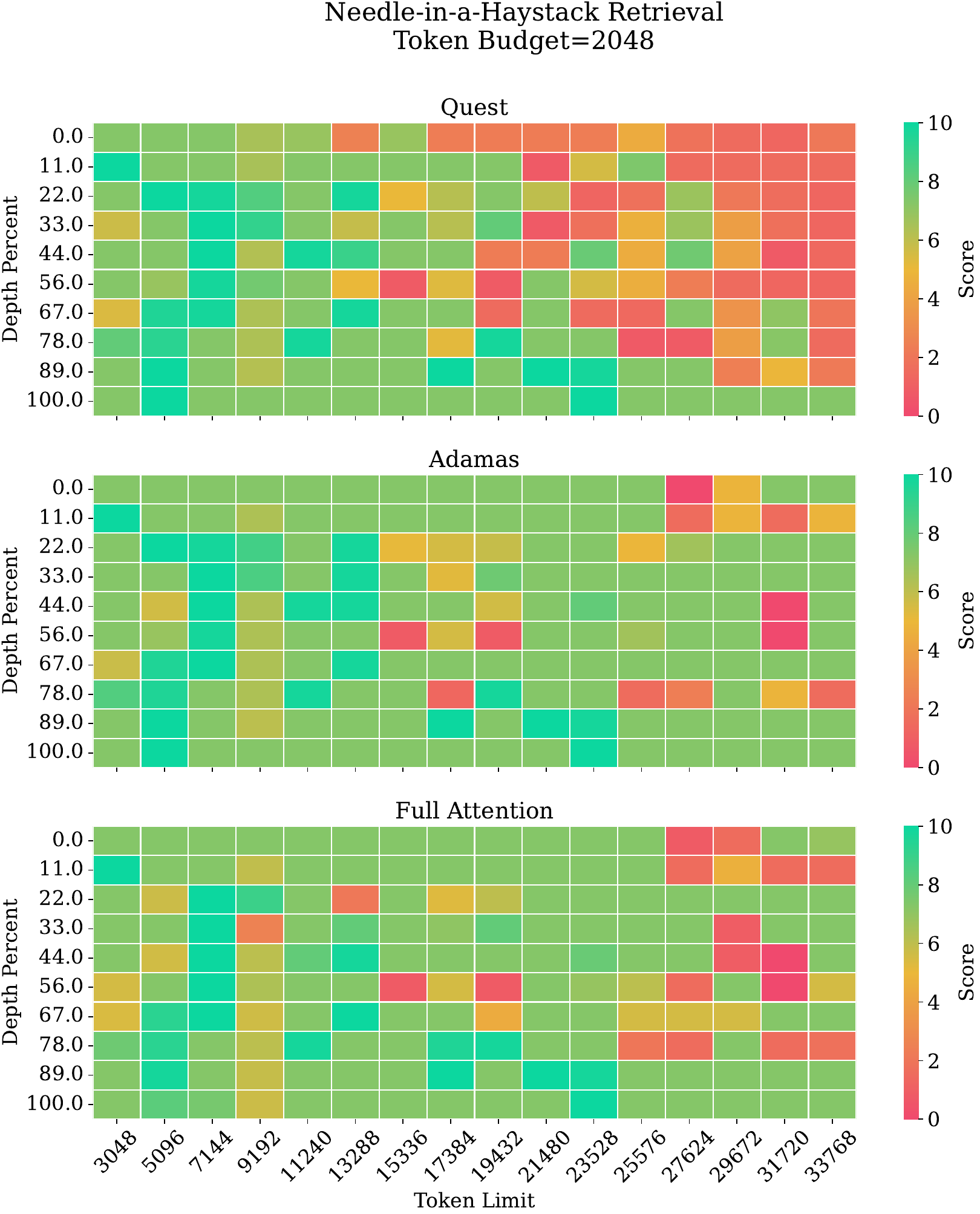}
    \end{subfigure}

    \caption{NIAH stress test on Adamas (ours), Quest and full attention, using Yarn-Llama-2-7b-128k. Token budgets range from 256 to 2048. Adamas consistently outperforms Quest in all budgets.}
    \label{fig:NIAH}
\end{figure}

We conduct the Needle-in-a-Haystack (NIAH) stress tests using Yarn-Llama-2-7b-128k model to evaluate the recall rates of Adamas, Quest, and full attention at a context length of 32k. As shown in Figure~\ref{fig:NIAH}, Adamas maintains a recall performance that is nearly identical to that of full attention even under high sparsity (with only a 256-token budget), consistently outperforming the baseline. Specifically, both Adamas and full attention achieve scores around 7/10 at a sequence length of 32k, whereas Quest reaches only a little over 2/10. These results further demonstrate the strong retrieval capability of Adamas under high sparsity.

\end{document}